\RequirePackage{fix-cm}
\documentclass[twocolumn]{svjour3}          
\smartqed  
\usepackage{graphicx}

\usepackage{epsfig}
\usepackage{color}
\usepackage{epstopdf}
\usepackage{graphicx}
\usepackage{amsmath}
\usepackage{amsfonts}
\usepackage{amssymb}
\usepackage{mathrsfs}
\usepackage{cite}
\usepackage{url}
\usepackage{subfigure}
\usepackage{array}
\usepackage{xcolor}
\usepackage{caption}
\usepackage{booktabs}
\usepackage{algorithm}
\usepackage{algorithmic}
\usepackage{indentfirst}
\usepackage{calc}
\usepackage{eqparbox}
\usepackage{pifont}
\usepackage{makecell}
\usepackage{multirow}
\usepackage{setspace}
\usepackage{booktabs}
\usepackage{tabularx}
\usepackage{diagbox}
\usepackage{marvosym} 
\usepackage[pagebackref=true,colorlinks,bookmarks=false]{hyperref}
\usepackage{mathptmx}      
\providecommand{\bmhead}[1]{%
  \par\medskip\noindent{\bfseries #1}\hspace{0.8em}%
}
\usepackage{latexsym}
\usepackage{microtype}
\makeatletter

\renewcommand\subsection{\@startsection{subsection}{2}{\z@}%
  {-3.25ex\@plus -1ex \@minus -.2ex}%
  {1.5ex \@plus .2ex}%
  {\normalfont\normalsize\bfseries}}
\makeatother
\begin{document}
\title{CAMotion: A High-Quality Benchmark for Camouflaged Moving Object Detection in the Wild
}






\author{
Siyuan Yao$^{1}$ \thanks{Siyuan Yao and Hao Sun contributed equally to this work.}\and
Hao Sun$^{1}$ \and
Ruiqi Yu$^{2}$ \and
Xiwei Jiang$^{3}$ \and
Wenqi Ren$^{1}$ \and
Xiaochun Cao$^{1}$
}

\authorrunning{S. Yao et al.}

\institute{
\Letter\ Siyuan Yao \at
\email{yaosiyuan04@gmail.com}
\and 
Hao Sun \at
\email{haosun.academic@gmail.com}
\and 
Ruiqi Yu \at
\email{yuruiqi422@gmail.com}
\and 
Xiwei Jiang \at
\email{xwjiang888@gmail.com}
\and 
Wenqi Ren \at
\email{rwq.renwenqi@gmail.com}
\and 
Xiaochun Cao \at
\email{caoxiaochun@mail.sysu.edu.cn}
\and 
$^{1}$ School of Cyber Science and Technology, Shenzhen Campus of Sun Yat-sen University, Shenzhen, China \\[2mm]
$^{2}$ College of Computing and Data Science, Nanyang Technological University, Singapore \\[2mm]
$^{3}$ School of Computer Science (National Pilot Software Engineering School), Beijing University of Posts and Telecommunications, Beijing, China
}

\date{Received: date / Accepted: date}

\maketitle


\begin{abstract} 
Discovering camouflaged objects is a challenging task in computer vision due to the high similarity between camouflaged objects and their surroundings. While the problem of camouflaged object detection over sequential video frames has received increasing attention, the scale and diversity of existing video camouflaged object detection (VCOD) datasets are greatly limited, which hinders the deeper analysis and broader evaluation of recent deep learning-based algorithms with data-hungry training strategy. To break this bottleneck, in this paper, we construct CAMotion, a high-quality benchmark covers a wide range of species for camouflaged moving object detection in the wild. CAMotion comprises various sequences with multiple challenging attributes such as uncertain edge, occlusion, motion blur, and shape complexity, etc. The sequence annotation details and statistical distribution are presented from various perspectives, allowing CAMotion to provide in-depth analyses on the camouflaged object's motion characteristics in different challenging scenarios. Additionally, we conduct a comprehensive evaluation of existing SOTA models on CAMotion, and discuss the major challenges in VCOD task. The benchmark is available at \url{https://www.camotion.focuslab.net.cn}, we hope that our CAMotion can lead to further advancements in the research community.








\end{abstract}

\begin{figure*}[!h]
\centering
\includegraphics[width=\textwidth]{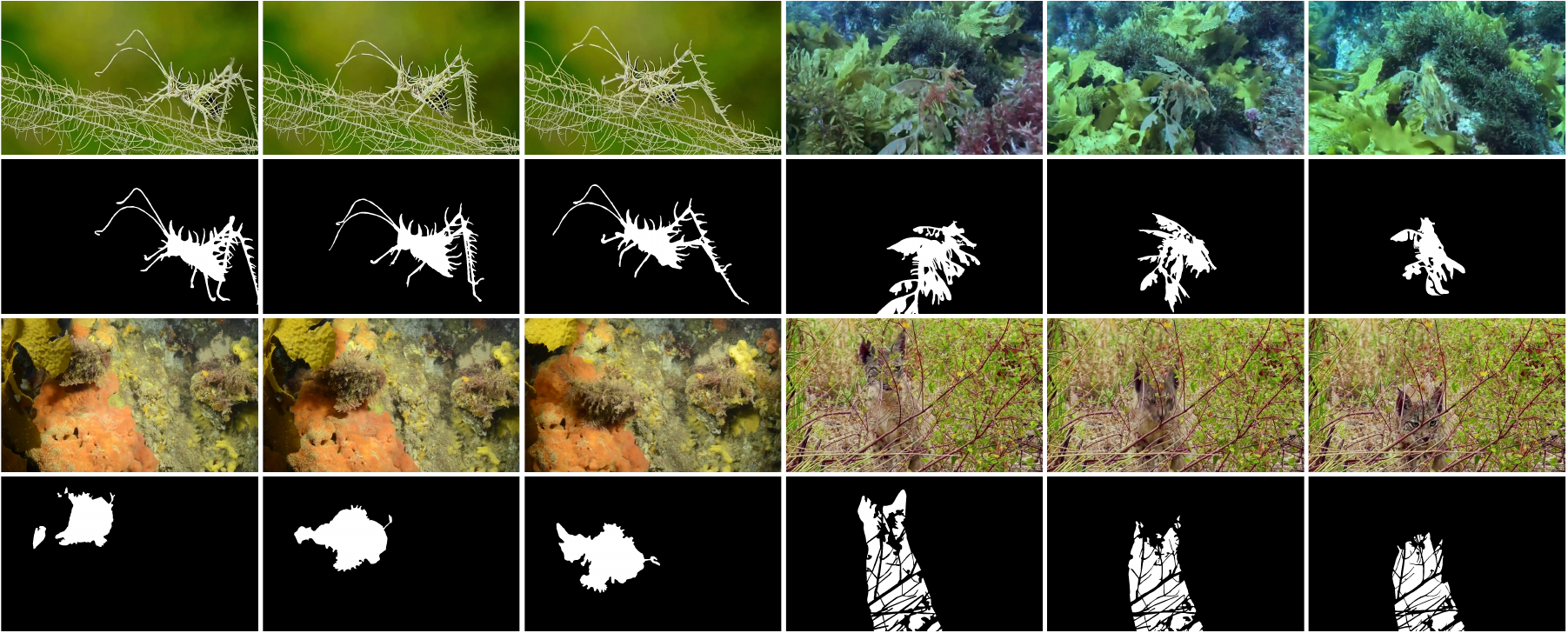}
\caption{Examples of our CAMotion dataset with corresponding pixel-level annotations. The first and third rows contain original images; the second and last rows contain corresponding pixel-wise ground truth annotations.}
\end{figure*}

\section{Introduction}

Camouflage is a widespread defensive behavior in natural scenarios that disguises the appearance to blend with the surroundings for deception and paralysis purposes. To distinguish the camouflaged objects in various challenging environments, Camouflaged Object Detection (COD) has become a prevalent topic in the computer vision community. Different from traditional dense prediction tasks, where objects typically exhibit distinct boundaries, camouflaged objects often share similar colors and textures with the background, making the objects difficult to perceive. This task becomes even more challenging for video sequences due to the dynamic appearance changes of objects and background over time.

\begin{table*}[h]
  \centering
  \footnotesize
  \caption{Statistics of camouflage datasets. $^{*}$ indicates that the \#Species is not reported in the original paper and is counted by us.} 
    \begin{tabular}{lcccccccc}
    \toprule
    \textbf{Dataset} & \textbf{Year}  & \textbf{Publication}  & \textbf{Type}  & \textbf{\#Img. }  & \textbf{\#Ann. Img.}  & \textbf{\#Species}  & \textbf{Bbox. GT} & \textbf{Mask GT} \\
    \midrule
    CAD\cite{DBLP:conf/eccv/BideauL16} & 2016 & ECCV &  Video  &   839    &  181    &  6   & \ding{55} & \ding{51} \\
    CHAMELEON\cite{skurowski2018animal}                                        & 2018 &  -   &  Image  &   76     &  76     &  27$^{*}$ & \ding{55} & \ding{51} \\
    CAMO\cite{le2019anabranch}                        & 2019 & CVIU &  Image  &  2,500   &  1,250  &  97$^{*}$ & \ding{55} & \ding{51} \\
    COD10K\cite{fan2020camouflaged}                   & 2020 & CVPR &  Image  &  10,000  &  5,066  &  69  & \ding{51} & \ding{51} \\
    MoCA\cite{DBLP:conf/accv/LamdouarYXZ20}               & 2020 & ACCV &  Video  &  37,250  &  7,617  &  67  & \ding{51} & \ding{55} \\
    CAMO++\cite{DBLP:journals/tip/LeCNLNDTN22}             & 2021 &  TIP &  Image  &  5,500   &  5,500  &  93  & \ding{51} &\ding{51} \\
    NC4K\cite{lv2021simultaneously}  & 2021 & CVPR &  Image  &   4,121  &  4,121  &  85$^{*}$ & \ding{55} & \ding{51} \\
    MoCA-Mask\cite{DBLP:conf/cvpr/ChengXFZHDG22}  & 2022 & CVPR &  Video  &  22,939  &  4,691  &  44  & \ding{55} & \ding{51} \\
    \midrule
    \textbf{CAMotion}       & 2026 &  -   &  Video  & 149,319  &  30,028 &  151 & \ding{51} & \ding{51} \\
    \bottomrule
    \end{tabular}%
  \label{tab:tab1}%
\end{table*}%

With the advent of deep learning-based techniques in recent years, various camouflaged object detection datasets have been established for comprehensive analyses. CAMO \cite{le2019anabranch} and CHAMELEON \cite{skurowski2018animal} make the early efforts to explore the camouflaged object segmentation problem and construct camouflaged objects dataset for benchmarking. The subsequent datasets, such as COD10K \cite{fan2020camouflaged} and NC4K \cite{lv2021simultaneously}, expand the diversity of species and scenarios across various attributes, thereby facilitating more comprehensive evaluation of concealed objects and advancing progress in relevant downstream vision tasks. Concurrently, some researchers also explore discovering the camouflaged objects in consecutive video sequences. The representative works like CAD \cite{DBLP:conf/eccv/BideauL16} and MoCA-Mask \cite{DBLP:conf/cvpr/ChengXFZHDG22} datasets provide pixel-wise annotations and conduct a preliminary investigation into the motion characteristics of camouflaged objects.  


Despite the research efforts,  several critical issues persist in the evaluation of video camouflaged object detection (VCOD) algorithms. First, although deep learning-based models have dominated the research field, the scale of existing VCOD dataset is greatly limited, which hinders to investigate the potential of recent deep learning-based algorithms with data-hungry training strategy. Second, since VCOD requires conducting pixel-wise camouflaged objects prediction in unconstrained environments, the data diversity is thus vital for fair evaluation. Nevertheless, existing VCOD datasets suffer from the limited scale of scenes and species. As a result, the generalization capabilities of existing VCOD algorithms are obscure. Moreover, as numerous attributes, e.g., complex shape and occlusion, may be involved in the video frames, the effectiveness of existing camouflaged object detectors in these challenging attributes is still unclear.

To address these issues, in this paper, we construct CAMotion, a high-quality benchmark covers a wide range of species for camouflaged moving object detection in the wild. CAMotion consists of approximately 150K video frames categorized into 151 species, in which 30,028 frames have been carefully annotated. The major properties of CAMotion are summarized as follows.


\begin{enumerate}
    \item{\textbf{Large-Scale}. The CAMotion dataset collects approximately 150K frames across 474 videos, which significantly exceeds the existing largest VCOD dataset by more than six times in terms of frame numbers. The videos within CAMotion present complicated challenges that necessitate a more robust VCOD model to effectively tackle and decipher them.}
    \item{\textbf{Diverse Categories and Species}. The constructed CAMotion dataset follows a biology-inspired hierarchical categorization. The video sequences span 12 classes that can be further classified into 50 subclasses and 151 species. These species are distributed in a wide range of regions and ecosystems, including terrestrial, aerial, and aquatic, ensuring environmental diversity around the world.}
    \item{\textbf{High-Quality Annotations}. The frames in the CAMotion dataset have been manually and precisely annotated through a multi-round feedback process. We provide both mask and bounding box annotations at a five-frame interval for each sequence, encompassing a total of over 30,000 annotation frames. Each video sequence is also carefully labeled with eight attributes, providing abundant samples for in-depth analyses across various challenge scenarios.}
\end{enumerate}

We conduct comprehensive experiments on the CAMotion dataset to evaluate the performance of 18 COD/VCOD models. Despite the promising performance of these models on existing datasets, these models suffer from a notable performance decline in the CAMotion benchmark. Either the COD or VCOD methods struggle to balance the camouflaged discriminative capability and temporal consistency. How to accurately identify camouflaged objects through video frames while alleviating the error accumulation over time is a crucial challenge. Compared to another VCOD dataset, MoCA-Mask, CAMotion exhibits more stable evaluation results and well-balanced camouflaged objects' diversity. Through attribute-based analysis and visualization of prediction results, we discover that the major challenges stem from small object (SO), uncertainty edge (UE), occlusion (OC) and multiple objects (MO). Additionally, we analyze the class-based performance and motion patterns of the camouflaged objects, aiming to uncover the root causes of the unsatisfactory performance and illuminate potential avenues for future improvement. 	


In conclusion, the contributions of this paper are summarized below:

$\bullet$ We construct a high-quality VCOD dataset, CAMotion, which comprises various sequences with multiple challenging attributes and a wide range of species for camouflaged moving object detection in the wild. 

$\bullet$ We present annotation details and statistical distribution of the collected dataset from various perspectives, allowing CAMotion to provide in-depth analyses on the camouflaged object's motion characteristics in different challenging scenarios.

$\bullet$ We conduct a comprehensive evaluation on the CAMotion dataset using recent SOTA COD/VCOD models, and reveal the major challenges in the VCOD task.

\section{Related Work}
\label{sec:formatting}

\subsection{Camouflaged Object Detection} 

Camouflaged object detection (COD) aims to discover camouflaged objects from a single RGB image. Inspired by the concealment strategy in biology, some approaches \cite{fan2020camouflaged, sun2021context, zhang2022preynet, DBLP:journals/pami/PangZXZL24, DBLP:journals/ijcv/YinZLCLH25} simulate the behavior process of predators to search and locate camouflaged objects. For example, SINet \cite{fan2020camouflaged} utilizes a searching module and an identification module to locate and detect objects with similar background distractions. ZoomNet \cite{pang2022zoom} imitates human vision by zooming in and out the imperceptible camouflaged objects with mixed scales. Another strategy is the multi-task joint learning-based approach \cite{li2021uncertainty, zhai2021mutual, DBLP:journals/tip/LiYZWZQ22, he2023camouflaged, he2023strategic, 10132418, DBLP:journals/tip/ZhouZGYZ22, DBLP:journals/tip/HaoYLXYY25, Ye_2025_ICCV}. These methods typically utilize auxiliary tasks to segment the camouflaged objects. For instance, in \cite{DBLP:journals/tip/LiYZWZQ22,he2023camouflaged,he2023strategic, Ye_2025_ICCV}, the boundary-aware priors are introduced to extract features that highlight the structural details of the object. \cite{li2021uncertainty} and \cite{DBLP:journals/tip/HaoYLXYY25} propose general segmentation models that jointly address the detection task of salient and camouflaged objects. Besides, PUENet \cite{10159663} models epistemic uncertainty and aleatoric uncertainty for effective segmentation with less model and data bias. \cite{zhong2022detecting, DBLP:conf/eccv/SunXYXL24, DBLP:journals/tip/JiXPZS25} introduce visual cues in the frequency domain to capture the subtle details of camouflaged objects from the background. \cite{wu2023source, Liu_2025_ICCV} attempt to utilize the complementary information in the depth map to assist in detection. With the growing attention paid to diffusion models, FocusDiffuser \cite{DBLP:conf/eccv/ZhaoLYZLJC24} and CamoDiffusion \cite{DBLP:journals/pami/SunCLSLJ25} introduce a new learning paradigm that employs a conditional diffusion model to generate masks that progressively refine the boundaries of camouflaged objects. \cite{DBLP:journals/tip/SongKWLLL25} first studies COD from a continuous feature representation perspective, transforming hierarchical features into a continuous function for the discovery of subtle discriminative clues. To further improve training efficiency, \cite{Sun_2025_ICCV} leverages the MoE strategy to adaptively modulate frozen foundation models to adapt the COD task.

Due to the intrinsic similarity of camouflaged objects, annotating camouflaged objects pixel-wise is very time-consuming and labor-intensive. To alleviate the heavy annotation burden, \cite{DBLP:conf/aaai/HeDLL23} proposes the first weakly-supervised COD dataset with scribble annotation and utilizes low-level contrasts to locate camouflaged objects. \cite{DBLP:conf/nips/HeLZXTZGL23, DBLP:conf/eccv/ChenWGG24, DBLP:journals/pami/HeLZYPTFZKLF25} present novel unified frameworks inheriting from SAM, integrating scribble, bounding box, and point for weakly-supervised camouflaged object detection. \cite{DBLP:conf/eccv/ChenSGG24} proposes the first point-supervised COD dataset and develops a COD method by imitating the cognitive process of the human vision system under the guidance of point supervision. \cite{DBLP:conf/eccv/ZhangZSCLK24} introduces a noise correction loss to correct pseudo labels with seriously noisy pixels. Furthermore, researchers also explore the semi-supervised \cite{DBLP:journals/pami/HeLZYPTFZKLF25, DBLP:conf/eccv/LaiYHZCJWZJ24}, unsupervised \cite{DBLP:conf/cvpr/YanCKZZC25, DBLP:conf/cvpr/DuHYKWWXL25, Du_2025_ICCV}, and zero-shot \cite{DBLP:journals/tip/LiFXZYC23, DBLP:journals/tip/DuWKLHXWWL25, DBLP:journals/pami/LeiFLXLZZ25} COD,  which helps to mitigate the intensive annotation cost.

\subsection{Video Camouflaged Object Detection} 

In contrast to static COD tasks, video camouflaged object detection (VCOD) leverages both appearance and temporal information between video frames to break camouflage. Early works \cite{DBLP:conf/accv/LamdouarYXZ20, DBLP:conf/iccv/YangLLZX21, DBLP:conf/nips/XieXZ22, DBLP:journals/pami/MeunierBB23, DBLP:journals/ijcv/BideauLSA24} handle VCOD as a motion segmentation problem which utilizes the predicted optical flow to explicitly model the spatio-temporal correlation between frames. Cheng \emph{et~al.} \cite{DBLP:conf/cvpr/ChengXFZHDG22} proposes a transformer-based model to implicitly model both short and long-term temporal consistency between frames. Besides, they propose MoCA-Mask, a dataset which selects 87 camouflaged video sequences from MoCA with pixel-level handcrafted labeling. ZoomNeXt \cite{DBLP:journals/pami/PangZXZL24} imitates human vision by zooming in and out video frames to perceive camouflaged objects and utilizes temporal shift to propagate inter-frame differences. EMIP \cite{DBLP:journals/tip/ZhangXJWFZ25} explicitly handles motion cues via a frozen pre-trained optical flow fundamental model. VSCode \cite{DBLP:conf/cvpr/LuoLZYZFKH24} and VSCode-v2 \cite{DBLP:journals/pami/LuoLYZFKH26} propose generalist models for multimodal binary segmentation tasks, taking RGB image and optical flow as input to perform frame-by-frame camouflaged object discovery across videos. With the emergence of visual foundation models, several methods \cite{DBLP:conf/cvpr/HuiZ0024, DBLP:conf/cvpr/MeeranTM22} take advantage of the exceptional segmentation performance of SAM \cite{DBLP:conf/iccv/KirillovMRMRGXW23} to segment camouflaged objects in videos by injecting temporal information into the prompt and SAM features. CamSAM2 \cite{zhou2025camsam} further leverages the strong generalizability in natural videos of SAM2 \cite{DBLP:conf/iclr/RaviGHHR0KRRGMP25} to address the VCOD task. However, due to the limitation posed by the low diversity of MoCA-Mask, most VCOD methods require pre-training on image datasets, e.g., COD10K, and more importantly, this constraint impedes the further advancement of this task. Although recent efforts~\cite{DBLP:journals/ijcv/VuZSGTY26} expand the number of annotated video frames, a large-scale VCOD dataset with high-quality annotations is still needed for comprehensive evaluation of existing methods.

\begin{figure*}[!t]
\centering
\includegraphics[width=6.5in]{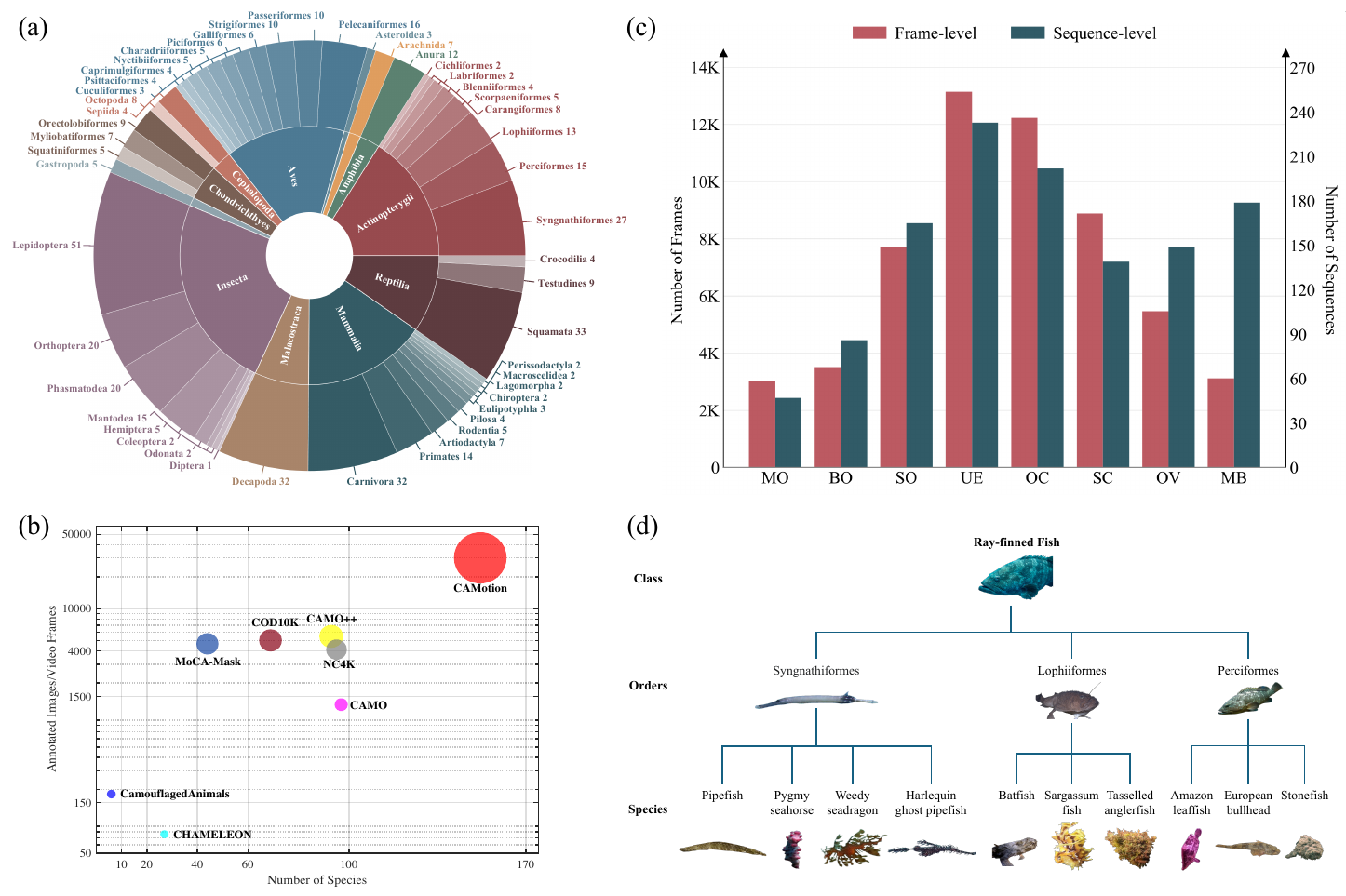}
\caption{Dataset features and species examples from CAMotion dataset. (a) Taxonomic structure of CAMotion. (b) The scale and species comparison between existing COD dataset and CAMotion. (c) The attributes distribution in frame-level and sequence-level. (d) Examples of the \textit{Ray-finned Fish} class in CAMotion. Please zoom in for details.}
\label{fig:illustration}
\end{figure*}

\begin{table*}[t]
    \centering
    \footnotesize
    \caption{Comparison of COD, VCOD, and related segmentation tasks.}
    \label{tab:task_settings}
    
    \begingroup
    \setlength{\tabcolsep}{4pt}
    \renewcommand{\arraystretch}{1.18}
    
    \begin{tabularx}{\textwidth}{
    @{}
    >{\raggedright\arraybackslash}p{0.16\textwidth}
    >{\raggedright\arraybackslash}p{0.27\textwidth}
    >{\raggedright\arraybackslash}p{0.21\textwidth}
    >{\raggedright\arraybackslash}X
    @{}}
    \toprule
    \textbf{Task} &
    \textbf{Input} &
    \textbf{First-frame prompt} &
    \textbf{Evaluation metrics} \\
    \midrule
    COD
    & Single image
    & None
    & $S_{\alpha}$, $F_{\beta}^{w}$, $E_{\phi}^{m}$, $\mathit{M}$ \\
    \midrule
    Semi-supervised VOS
    & Video sequence
    & Mask
    & $\mathit{J}$, $\mathit{F}$, $\mathit{J}\&\mathit{F}$ \\
    \midrule
    PVS 
    & Video sequence
    & Point, bounding box, or mask
    & $\mathit{J}\&\mathit{F}$ \\
    \midrule
    Motion segmentation
    & Motion cues derived from video frames
    & None
    & $\mathit{J}$ \\
    \midrule
    VCOD
    & Video clip or sequence
    & None
    & $S_{\alpha}$, $F_{\beta}^{w}$, $E_{\phi}^{m}$, $\mathit{M}$,
    $\mathrm{mDic}$, $\mathrm{mIoU}$, $\mathbf{dtSSD}$~\cite{DBLP:conf/bmvc/ErofeevGVFW15} \\
    \bottomrule
    \end{tabularx}
    \endgroup
\end{table*}

\subsection{Motion Segmentation}

Motion segmentation is a fundamental task in computer vision that aims to partition a video sequence into regions based on their motion characteristics. By prioritizing movement over visual appearance, it provides a powerful mechanism to address challenging scenarios where standard visual cues is insufficient, such as fast motion, occlusion, deformation, and low contrast scenarios. Existing approaches can be broadly categorized into two dominant paradigms: flow-based methods, which focus on short-term, dense motion cues, and trajectory-based methods, which model long-term, sparse motion patterns. For flow-based methods, the early researches \cite{DBLP:conf/eccv/BideauL16, DBLP:conf/iccv/PapazoglouF13} perform object segmentation by manually grouping motion cues derived from optical flow. Recently, numerous deep learning-based approaches \cite{DBLP:journals/pami/HuWKKT20, DBLP:conf/wacv/FaisalAAH20, DBLP:journals/tip/PanDLPP20, DBLP:journals/tip/ZhuoC0WK20, DBLP:journals/tip/YangHNHMW22, DBLP:conf/eccv/XieXZ24, DBLP:conf/accv/XieYXZ24} leverage CNNs or attention mechanisms to extract motion cues from optical flow. For example, \cite{DBLP:conf/eccv/XieXZ24} introduce an appearance-based refinement method that leverages temporal consistency in video streams to correct inaccurate flow-based proposals. \cite{DBLP:conf/accv/XieYXZ24} leverages SAM to capture motion cues from optical flow, and uses the flow as input prompts. Besides, \cite{DBLP:journals/pami/MeunierB26} takes as input the volume of consecutive optical flow fields, and delivers a volume of segments of coherent motion over the video. Despite their effectiveness in capturing motion cues, flow–based methods often struggle with complex multi-object motions, and the short-term nature limits the ability of flow–based methods to handle long-term or occlusion movements.

Another widely adopted paradigm, trajectory-based methods, aims to overcome these limitations by modeling long-term, coherent motion patterns across frames. These methods conduct motion segmentation by applying geometrical constraints to motion subspaces \cite{DBLP:conf/eccv/YanP06, DBLP:conf/cvpr/RaoTVM08, DBLP:conf/nips/KarazijaL0V24} or employing non-negative matrix factorization algorithm \cite{DBLP:conf/iccv/CheriyadatR09}. Several works construct graphs over trajectories, employing specialized solvers for optimization \cite{DBLP:conf/iccv/KeuperAB15, DBLP:conf/iccv/Keuper17} or utilizing spectral clustering on hypergraphs to group trajectories into coherent motion segments \cite{DBLP:conf/cvpr/OchsB12, DBLP:journals/pami/OchsMB14}. Most recent work \cite{DBLP:conf/cvpr/HuangZXKZKW25} combines long-range trajectory motion cues with DINO-based semantic features and leverages SAM2 for pixel-level mask densification through an iterative prompting strategy. While effectively modeling long-term trajectory affinities, trajectory-based methods struggle with dynamic motion patterns and global consistency.

\section{CAMotion Dataset}
\label{sec:method}

\subsection{Task Definition}
\label{sec:task_definition}

CAMotion is constructed for video camouflaged object detection (VCOD). To distinguish VCOD from related tasks, we summarize their definitions as follows:

\begin{itemize}
    \renewcommand{\labelitemi}{\textbullet}
    \item \textbf{Camouflaged Object Detection (COD)}~\cite{fan2020camouflaged}
    aims to identify and segment objects that seamlessly blend into their surroundings in a single image.

    \item \textbf{Semi-supervised Video Object Segmentation (VOS)}~\cite{DBLP:conf/cvpr/PerazziPMGGS16,DBLP:conf/iccv/ChengTW017}
    aims to segment specified target objects in subsequent video frames, given their ground-truth masks in the first frame.

    \item \textbf{Promptable Visual Segmentation (PVS)}~\cite{DBLP:conf/iclr/RaviGHHR0KRRGMP25}
    aims to segment regions specified by prompts in images or videos. In the video setting, the first frame point, bounding box, or mask prompts can be provided to specify the target.

    \item \textbf{Motion Segmentation}~\cite{DBLP:conf/eccv/BideauL16,DBLP:conf/accv/LamdouarYXZ20}
    aims to segment moving objects from the background using motion cues, \emph{e.g.}, optical flow.

    \item \textbf{Video Camouflaged Object Detection (VCOD)}~\cite{DBLP:conf/cvpr/ChengXFZHDG22},
    also termed camouflaged moving object detection, aims to segment objects that seamlessly blend into their surroundings in video sequences.
\end{itemize}

Table~\ref{tab:task_settings} summarizes the differences of their inputs, target prompts, and evaluation metrics. Our CAMotion follows the VCOD setting, where camouflaged objects are segmented from video sequences without first-frame prompts. Similar to MoCA-Mask~\cite{DBLP:conf/cvpr/ChengXFZHDG22}, CAMotion provides pixel-level annotations at a 5-frame interval. Only annotated frames are used as inputs during training and inference.

\subsection{Video Collection}
The limited scale of existing VCOD dataset seriously hinders the comprehensive evaluation of recent VCOD algorithms. To address this issue, we build a large-scale VCOD dataset CAMotion with high-quality pixel-wise annotations. The whole dataset is collected from the viewpoint of biology-inspired hierarchical categorization. We retrieve from the Internet using the keywords \emph{camouflaged mammals}, \emph{concealed insects}, \emph{camouflaged fishes}, etc. Consequently, we obtain 151 representative camouflaged species, which significantly enriches the diversity of existing VCOD datasets with less than 50 species. Details of the camouflaged object classes and species can be found in Appendix B.1.

After determining the biology-inspired species, we collect more than 4,000 videos as the initial camouflaged videos. Then we evaluate the quality of these camouflaged videos, filter out the unrelated contents in each video, and retain the usable clip containing camouflaged objects. As a result, we construct CAMotion, comprising 474 video sequences with around 150K video frames. We split the total of 474 sequences into 359 sequences as training set and the other 115 sequences as testing set. In this dataset, the length of the video sequences varies from 114 frames to 1,063 frames. Similar to MoCA-Mask, we provide both mask and bounding box annotations with an interval of five frames per sequence, accounting for 30,028 annotation frames in total.


\subsection{Sequence Annotation}

The quality of the annotation plays a crucial role in the dense prediction task. To this end, we present high-quality pixel-wise annotation in CAMotion, which is significantly larger than existing COD datasets, \emph{e.g.} COD10K, NC4K, and VCOD dataset, \emph{e.g.} MoCA-Mask.


\begin{figure*}[!t]
\centering
\includegraphics[width=\textwidth]{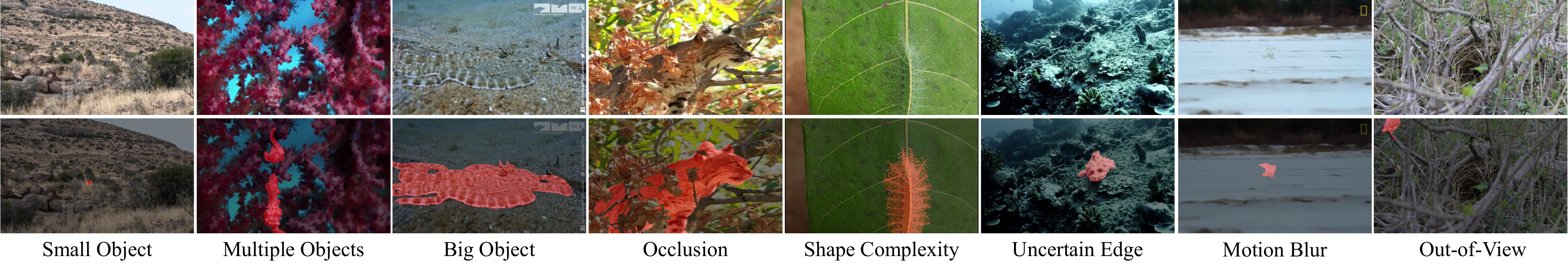}
\caption{Visualization of the challenging attributes in CAMotion. Best viewed in color and zoomed in for details.}
\label{fig:attribute_show}
\end{figure*}


\begin{figure}[t]
    \centering
    \includegraphics[width=\linewidth]{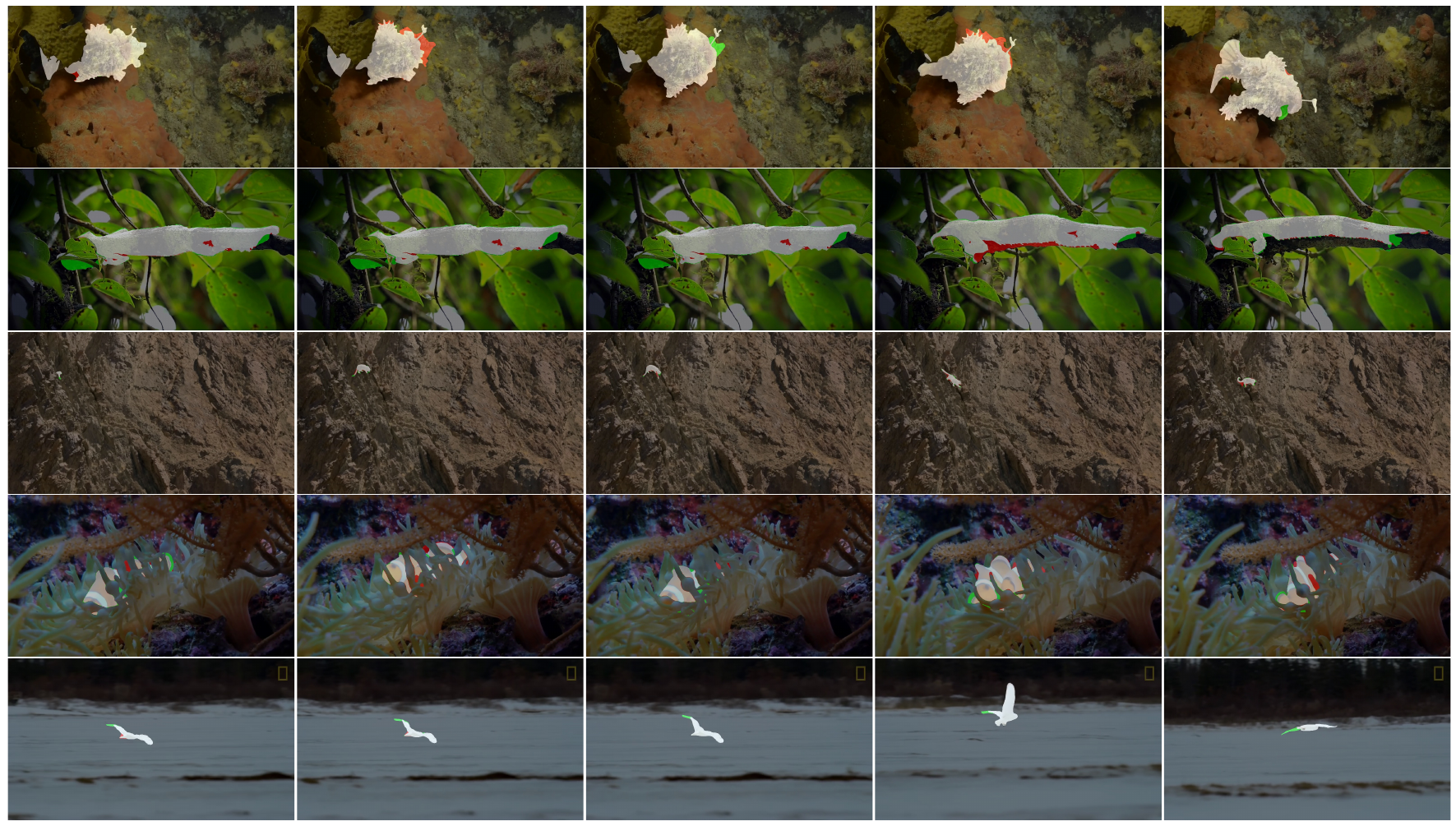}
    \caption{Examples of refined initial annotations. White denotes unchanged regions; red and green indicate over-annotated and previously missing regions in the original annotations, respectively. Please zoom in for details.}
    \label{fig:annotation}
\end{figure}


\begin{figure}[t]
    \centering
    \includegraphics[width=\linewidth]{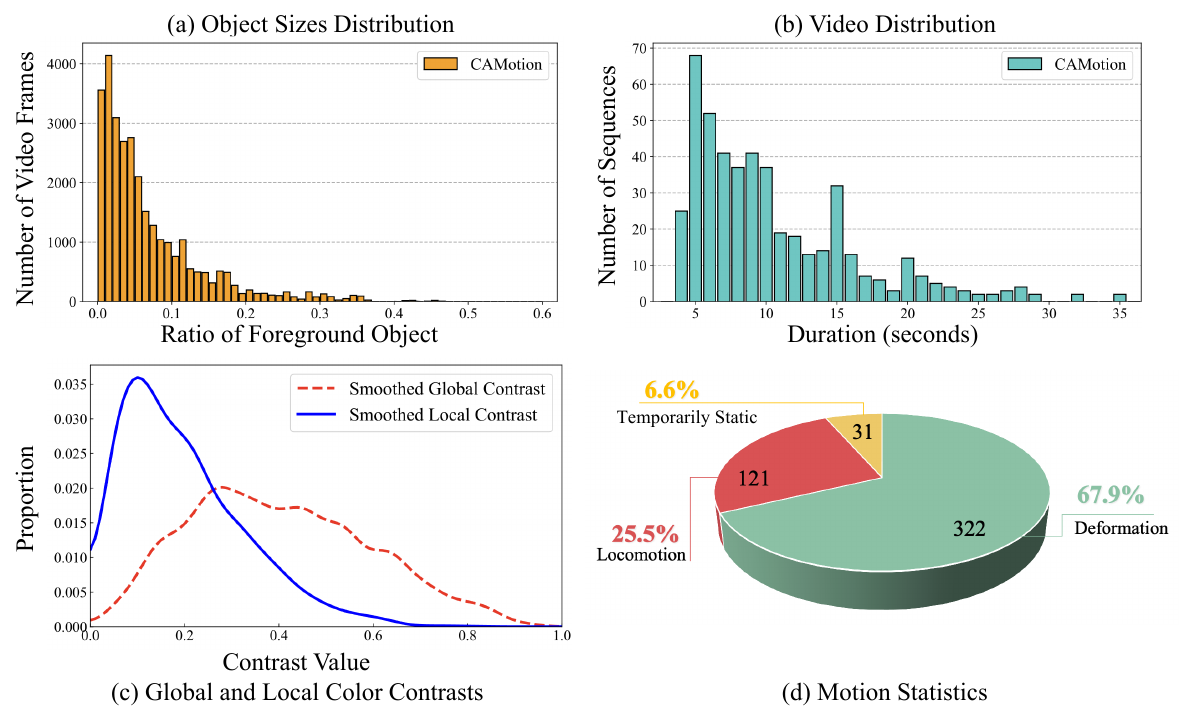}
    \caption{Statistics for CAMotion dataset. (a) Object sizes distribution. (b) The distribution of video durations. (c) Global/local contrast distribution. (d) Motion statistics of the camouflaged objects.}
    \label{fig:statistics}
\end{figure}


\textbf{Classes and species.} As shown in Fig. \ref{fig:illustration} (a), the camouflaged videos in our dataset follow a biology-inspired hierarchical categorization. All of the video sequences are firstly divided into 12 classes, including \emph{mammals, insects, birds, ray-finned fish, etc}. Then these videos are further classified into 50 subclasses, which can be regarded as the biological orders like \emph{carnivora, primates, and lepidoptera, etc}. To present more detailed analyses, we further categorize these data into 151 species, such as \emph{polar bears, dragonflies, tigers, cats, and batfishes, etc}. A representative taxonomic hierarchy tree of \emph{Ray-finned Fish} is demonstrated in Fig. \ref{fig:illustration} (d). To the best of our knowledge, our CAMotion is the largest VCOD dataset with diverse species in the research community. 

\textbf{Attributes.} To present deep analyses of the camouflaged videos in various challenging scenes, we label each camouflaged video with eight attributes, including uncertain edge (UE), big object (BO), multiple objects (MO), small object (SO), occlusions (OC), shape complexity (SC), out-of-view (OV) and motion blur (MB). The details of each attribute description are provided in Table \ref{tab:attr}. We provide attribute annotations for all the video frames in our dataset.

From Fig. \ref{fig:illustration} (c), we observe that the most common challenge factors in CAMotion are uncertain edge (UE), occlusions (OC), shape complexity (SC), and small object (SO). Such observations align with the intuitive reality that camouflaged objects in the local region are seamlessly blended into the surrounding backgrounds, thereby making the camouflaged objects imperceptible in these challenging scenes. Compared to MoCA \cite{DBLP:conf/accv/LamdouarYXZ20} and MoCA-Mask \cite{DBLP:conf/cvpr/ChengXFZHDG22} that simply categorized into three types of motion, \emph{i.e.} static, locomotion, and deformation, CAMotion can provide more comprehensive attributes for camouflaged behavior analyses. The representative examples of the challenging attributes in CAMotion are presented in Fig.~\ref{fig:attribute_show}.


\begin{table}[h]
    \centering
    \footnotesize
    \caption{List and description of the eight attributes that characterize videos in CAMotion.}
    \begin{tabularx}{\columnwidth}{lX}
        \toprule
        \textbf{Attr} & \textbf{Description} \\
        \midrule
        \textbf{MO}  & \underline{Multiple Objects}: image contains at least two objects.\\
        \textbf{BO}  & \underline{Big Object}: ratio between object area and image area $\ge$0.15.\\
        \textbf{SO}  & \underline{Small Object}: ratio between object area and image area $\le$0.02.\\
        \textbf{UE}  & \underline{Uncertain Edge}: the foreground and background areas around object have similar colors and textures.\\
        \textbf{OC} & \underline{Occlusion}: the object is partially occluded. \\
        \textbf{SC}  & \underline{Shape Complexity}: object contains thin parts (e.g., animal foot). \\
        \textbf{OV}  & \underline{Out-of-View}: some portion of the object leaves the camera field of view. \\
        \textbf{MB}  & \underline{Motion Blur}: the object region is blurred due to the motion of object or camera. \\
        \bottomrule
    \end{tabularx}
    \label{tab:attr}
\end{table}


\textbf{Quality control.} We make great effort to present precise annotations on the collected videos, and conduct feedback error correction to ensure the annotation quality. Specifically, we ask five annotators to identify the camouflaged instances in each image and use an interactive segmentation tool to annotate them via pixel-wise masks. It takes each annotator 3 to 20 minutes to annotate an image depending on its complexity. The annotator manually draws/edits the camouflaged object's boundary in each frame, and two other annotators inspect the results and adjust them if necessary. Afterwards, the annotation results are reviewed by two experts with professional knowledge on VCOD task. If an annotation result is not unanimously agreed by the experts, it will be sent back to the original annotators to revise. To improve the annotation quality as much as possible, the annotators are required to annotate these challenging video frames very carefully and revise them frequently. More than $60\%$ of the initial annotations fail in the first round of validation. Some crucial video frames are revised more than three times. We present some challenging frames that are initially labeled inaccurately in Fig. \ref{fig:annotation}. With all these efforts, we finally construct CAMotion dataset with high-quality dense annotation.

\subsection{Dataset Specification and Statistics}

\textbf{Object size.} Fig. \ref{fig:statistics} (a) illustrates the object size distribution in the proposed CAMotion dataset, where the reported ratio is defined as the proportion of foreground area relative to the entire image. The distribution is heavily skewed towards smaller dimensions, with the majority of object sizes falling within the 0.01 to 0.1 range, indicating the dataset is rich in tiny and small camouflaged objects. This is a critical feature for benchmarking VCOD methods, as detecting such minuscule and well-concealed objects remains a persistent difficulty for recent state-of-the-art models. Furthermore, CAMotion also contains a certain number of camouflaged objects with sizes ranging from $\left [ 0.1,0.35 \right ] $, ensuring a diverse size representation. This breadth makes the dataset well-suited for providing comprehensive and robust analyses on how object size impacts the performance of VCOD algorithms.

\textbf{Duration.} To evaluate the temporal adaptability of the COD/VCOD algorithms, we ensure that each sequence in CAMotion comprises at least four seconds with more than 114 frames, establishing a solid baseline for analyzing short-term motion patterns. The average sequence length in CAMotion is around 315 frames (see Fig. \ref{fig:statistics} (b)), which is substantially longer than existing benchmarks. To further evaluate the long-term dependency modeling, the dataset includes challenging videos that persist for nearly 35 seconds and contain more than 1,000 frames in a single clip. Consequently, the video durations in CAMotion are not only longer on average but also offer a greater range of temporal complexity compared to the previous MoCA-Mask dataset. The extended duration is critical for benchmarking advanced capabilities such as long-term object persistence, robustness to temporary full occlusions, and the stability of predictions against complex background motion and camouflage degradation over time.


\textbf{Global and local contrast.} We adopt global and local color contrast distributions to measure the detection difficulty of camouflaged objects in CAMotion dataset. As shown in Fig. \ref{fig:statistics} (c), the camouflaged objects in most video frames exhibit remarkably low local contrast. This indicates a high degree of similarity between the objects and their immediate surroundings, making them exceptionally difficult to distinguish using local appearance cues alone. Conversely, the broader distribution of global contrast values indicates that CAMotion encompasses a wide range of species and scene diversity. Such low local contrast and broad global contrast make CAMotion a challenging and comprehensive benchmark for the VCOD task.

\textbf{Motion statistics.} Fig.~\ref{fig:statistics} (d) shows the motion statistics of the camouflaged objects in CAMotion dataset. The three motion labels, \emph{i.e.}, locomotion, deformation, and temporarily static, are assigned at the sequence level and are mutually exclusive, with each sequence assigned one label according to the following definitions:

\begin{itemize}
    \renewcommand{\labelitemi}{\textbullet}
    \item \textbf{Locomotion}: The target exhibits movements that substantially change its location.
    \item \textbf{Deformation}: The target changes its pose while remaining in approximately the same location.
    \item \textbf{Temporarily Static}: The target remains stationary during parts of the video, without noticeable changes in its location or pose.
\end{itemize}

To ensure high annotation quality, the labeling was conducted by five annotators following the same protocol adopted for mask annotation. A critical observation is that the overwhelming majority of objects ($93.4\%$) exhibit either locomotion or deformation, while only $6.6\%$ remain static without obvious motion or appearance changes. More importantly, compared to MoCA-Mask, the camouflaged objects in CAMotion demonstrate more complex and informative motion cues. This richness arises from the frequent camera pose variations, intricate body-part movements, and dynamic environmental factors (\emph{e.g.,} camouflaged insects on swaying petals). Such motion diversity ensures that the dataset includes a wider range of motion challenges, providing a more comprehensive benchmark for evaluating motion patterns of camouflaged objects.

\section{Experiment}
\label{sec:exp}



\subsection{Experiment Settings}

\textbf{Datasets.} We use two VCOD datasets, MoCA-Mask \cite{DBLP:conf/cvpr/ChengXFZHDG22} and our CAMotion, and an image COD dataset, COD10K \cite{fan2020camouflaged} to conduct the experiments. MoCA-Mask is reorganized from MoCA \cite{DBLP:conf/accv/LamdouarYXZ20}, which contains 71 sequences with 3,946 frames for training and 16 sequences with 745 frames for testing.
Our proposed CAMotion dataset includes 359 sequences with 23,253 frames for training and 115 sequences with 6,775 frames for testing. COD10K contains 3,040 training and 2,026 testing camouflaged images. Following the previous setting \cite{DBLP:conf/cvpr/ChengXFZHDG22, DBLP:journals/pami/PangZXZL24}, training conducted on MoCA-Mask is pretrained on COD10K and fine-tuned on MoCA-Mask. 

\textbf{Implementation details and metrics.} 
Given the diversity in network designs, input resolutions, modalities, and preprocessing strategies among baselines, we carefully follow the original settings specified in each method’s official implementation to ensure fair comparisons. We use input resolutions as the original setups: 352 $\times$ 352 for SegMaR~\cite{jia2022segment}, SINet-v2~\cite{fan2021concealed}, SLT-Net~\cite{DBLP:conf/cvpr/ChengXFZHDG22}, and EMIP~\cite{DBLP:journals/tip/ZhangXJWFZ25}; 384 $\times$ 384 for ZoomNet~\cite{pang2022zoom}, FSPNet~\cite{huang2023feature}, ZoomNeXt~\cite{DBLP:journals/pami/PangZXZL24}, CamoDiffusion~\cite{DBLP:journals/pami/SunCLSLJ25}, and RUN~\cite{DBLP:conf/icml/HeZX0T0KF0F25}; 416 $\times$ 416 for PFNet~\cite{mei2021camouflaged} and ESCNet~\cite{Ye_2025_ICCV}; 448 $\times$ 448 for DaCOD~\cite{DBLP:conf/mm/WangYYWCZ23}; 473 $\times$ 473 for MGL-R~\cite{zhai2021mutual} and UGTR~\cite{yang2021uncertainty}; 512 $\times$ 512 for PUENet~\cite{10159663}, PopNet~\cite{wu2023source}, and HGINet~\cite{DBLP:journals/tip/YaoSXWC24}; and 1024 $\times$ 1024 for SAM2~\cite{DBLP:conf/iclr/RaviGHHR0KRRGMP25}, SAM-PM~\cite{DBLP:conf/cvpr/MeeranTM22}, and CamSAM2~\cite{zhou2025camsam}. All experiments are conducted using four NVIDIA RTX L40 GPUs. Following \cite{DBLP:conf/cvpr/ChengXFZHDG22}, we use six common evaluation metrics, including S-measure ($S_{\alpha}$) \cite{Cheng2021sMeasure}, weighted F-measure ($F_{\beta}^{w}$) \cite{6909433}, mean E-measure ($E_{\phi}^{m}$) \cite{ijcai2018p97}, mean absolute error ($\mathit{M}$), mean Dice ($\mathrm{mDic}$), mean IoU ($\mathrm{mIoU}$), and add a temporal consistency metric $\mathbf{dtSSD}$ for CAMotion.



\begin{table*}[t]
  \setlength{\tabcolsep}{1.9pt}
  \footnotesize
  \centering
  \caption{Quantitative comparison with 20 cutting-edge methods on CAMotion and MoCA-Mask testing datasets. Notes $\uparrow/\downarrow$ denotes the higher/lower the better, and the best and second best are \textbf{bolded} and \underline{underlined} for highlighting, respectively. $^{\ddagger}$ indicates that the first-frame annotation is removed during both training and testing for fair comparison.}
    \begin{tabular}{llccccccccccccc}
    \toprule
    \multirow{2}[3]{*}{\textbf{Methods}} & \multirow{2}[3]{*}{\textbf{Publications}} &\multicolumn{7}{c}{\textbf{CAMotion}} & \multicolumn{6}{c}{\textbf{MoCA-Mask}} \\
    \cmidrule(lr){3-9} \cmidrule(lr){10-15} & & $S_{\alpha}\uparrow $ &  $F_{\beta}^{w}\uparrow $  & $E_{\phi}^{m}\uparrow $ & $\mathit{M}\downarrow$ & $\mathrm{mDic}\uparrow$ & $\mathrm{mIoU}\uparrow$ & $\mathbf{dtSSD}\downarrow$ & $S_{\alpha}\uparrow $ &  $F_{\beta}^{w}\uparrow $  & $E_{\phi}^{m}\uparrow $ & $\mathit{M}\downarrow$ & $\mathrm{mDic}\uparrow$ & $\mathrm{mIoU}\uparrow$\\
    
    \midrule
    MGL-R~\cite{zhai2021mutual}
      & CVPR'21
      & 0.542 & 0.176 & 0.604 & 0.078 & 0.195 & 0.129 & 0.106
      & 0.493 & 0.034 & 0.519 & 0.059 & 0.048 & 0.033 \\

    PFNet~\cite{mei2021camouflaged}
      & CVPR'21
      & 0.669 & 0.425 & 0.780 & 0.050 & 0.463 & 0.359 & 0.126
      & 0.558 & 0.142 & 0.633 & 0.026 & 0.172 & 0.118 \\

    UGTR~\cite{yang2021uncertainty}
      & ICCV'21
      & 0.687 & 0.403 & 0.720 & 0.048 & 0.440 & 0.342 & 0.106
      & 0.493 & 0.048 & 0.459 & 0.088 & 0.078 & 0.049 \\

    SegMaR\cite{jia2022segment}
      & CVPR'22
      & 0.645 & 0.377 & 0.695 & 0.049 & 0.402 & 0.311 & 0.112
      & 0.542 & 0.129 & 0.544 & 0.024 & 0.139 & 0.093 \\

    ZoomNet~\cite{pang2022zoom}
      & CVPR'22
      & 0.675 & 0.440 & 0.693 & 0.044 & 0.456 & 0.365 & 0.113
      & 0.582 & 0.201 & 0.682 & 0.026 & 0.236 & 0.197 \\

    SINet-v2~\cite{fan2021concealed}
      & TPAMI'22
      & 0.682 & 0.433 & 0.761 & 0.051 & 0.477 & 0.373 & 0.120
      & 0.571 & 0.175 & 0.608 & 0.035 & 0.211 & 0.153 \\

    FSPNet~\cite{huang2023feature}
      & CVPR'23
      & 0.725 & 0.515 & 0.759 & 0.037 & 0.535 & 0.437 & 0.103
      & 0.565 & 0.186 & 0.610 & 0.044 & 0.238 & 0.167 \\

    PUENet~\cite{10159663}
      & TIP'23
      & 0.744 & 0.562 & 0.842 & 0.041 & 0.607 & 0.493 & 0.133
      & 0.594 & 0.204 & 0.619 & 0.037 & 0.300 & 0.212 \\

    PopNet~\cite{wu2023source}
      & ICCV'23
      & 0.709 & 0.495 & 0.769 & 0.041 & 0.521 & 0.426 & 0.107
      & 0.613 & 0.317 & 0.694 & 0.035 & 0.307 & 0.219 \\

    DaCOD~\cite{DBLP:conf/mm/WangYYWCZ23}
      & ACM MM'23
      & 0.759 & 0.569 & 0.848 & 0.035 & 0.611 & 0.498 & 0.108
      &  0.605 & 0.235 & 0.620 & 0.021 & 0.274 & 0.204 \\

    HGINet~\cite{DBLP:journals/tip/YaoSXWC24}
      & TIP'24
      & \underline{0.774} & \textbf{0.634} & \underline{0.852}
      & \textbf{0.031} & \textbf{0.660} & \textbf{0.551} & 0.114
      & \underline{0.677} & \underline{0.403} & 0.744
      & \textbf{0.010} & \underline{0.441} & \underline{0.357} \\

    CamoDiffusion~\cite{DBLP:journals/pami/SunCLSLJ25}
      & TPAMI'25
      & 0.707 & 0.500 & 0.758 & 0.038 & 0.519 & 0.438 & 0.113
      & 0.676 & 0.382 & \underline{0.747} & \underline{0.012}
      & 0.410 & 0.340 \\

    RUN~\cite{DBLP:conf/icml/HeZX0T0KF0F25}
      & ICML'25
      & 0.711 & 0.500 & 0.792 & 0.048 & 0.540 & 0.433 & 0.125
      & 0.574 & 0.196 & 0.662 & 0.021 & 0.216 & 0.165 \\

    ESCNet~\cite{Ye_2025_ICCV}
      & ICCV'25
      & 0.718 & 0.525 & 0.781 & 0.039 & 0.552 & 0.455 & 0.115
      & 0.577 & 0.198 & 0.634 & 0.029 & 0.236 & 0.171 \\

    \midrule

    SAM2~\cite{DBLP:conf/iclr/RaviGHHR0KRRGMP25}$^{\ddagger}$
      & ICLR'25
      & 0.463 & 0.004 & 0.256 & 0.084 & 0.004 & 0.003 & \textbf{0.096}
      & 0.495 & 0.056 & 0.487 & 0.023 & 0.057 & 0.035 \\

    SLT-Net~\cite{DBLP:conf/cvpr/ChengXFZHDG22}
      & CVPR'22
      & 0.748 & 0.554 & 0.851 & 0.039 & 0.602 & 0.485 & 0.102
      & 0.631 & 0.311 & \textbf{0.759} & 0.027 & 0.360 & 0.272 \\

    SAM-PM~\cite{DBLP:conf/cvpr/MeeranTM22}$^{\ddagger}$
      & CVPRW'24
      & 0.741 & 0.550 & 0.802 & 0.047 & 0.589 & 0.488 & 0.101
      & 0.559 & 0.133 & 0.568 & 0.023 & 0.166 & 0.117 \\

    ZoomNeXt~\cite{DBLP:journals/pami/PangZXZL24}
      & TPAMI'24
      & \textbf{0.779} & \underline{0.593} & 0.832
      & \underline{0.033} & \underline{0.626} & \underline{0.523} & \underline{0.097}
      & \textbf{0.734} & \textbf{0.476} & 0.736
      & \textbf{0.010} & \textbf{0.497} & \textbf{0.422} \\

    EMIP~\cite{DBLP:journals/tip/ZhangXJWFZ25}
      & TIP'25
      & 0.761 & 0.583 & \textbf{0.866} & 0.035 & 0.617 & 0.506  & 0.105
      & 0.658 & 0.337 & 0.737 & 0.013 & 0.385 & 0.292 \\

    CamSAM2~\cite{zhou2025camsam}$^{\ddagger}$
      & NIPS'25
      & 0.626 & 0.378 & 0.701 & 0.075 & 0.393 & 0.328 & 0.116
      & 0.476 & 0.029 & 0.510 & 0.051 & 0.028 & 0.019 \\
    \bottomrule
    \end{tabular}%
    \label{tab:table3}%
\end{table*}


\begin{table*}[h]
    \footnotesize
    \centering
    \caption{$S_{\alpha}$ and $\mathit{M}$ results for cross-dataset generalization. 
    The selected ZoomNeXt is trained on one (rows) dataset and tested on all datasets (columns). ``Self'' refers to training and testing on the same dataset (same as diagonal), and ``Mean Others'' refers to averaging performance on all except self.}
    \begin{tabular}{clcccccc}
    \toprule
    \textbf{Metrics} & \diagbox{\textbf{Trained on}}{\textbf{Tested on}} & COD10K &  CAMotion  & MoCA-Mask & Self & \makecell{Mean \\ Others}  & \makecell{Performance\\Gap} \\
    \midrule
    \multirow{4}{*}{\textbf{$S_{\alpha}\uparrow$}}& COD10K & \textbf{0.897} & 0.836 & 0.686 & 0.897 & 0.761 & 0.136 \\
    & CAMotion & 0.832 & \textbf{0.774} & 0.690 & 0.774 & 0.761 & 0.013 \\
    & MoCA-Mask & 0.786 & 0.720 & \textbf{0.652} & 0.652 & 0.753 & 0.101  \\
    \cmidrule(lr){2-8}
    & Mean others & 0.809 & 0.778 & 0.688 & 0.774 & 0.758 & 0.016 \\
    \midrule
    \multirow{4}{*}{\textbf{$\mathit{M}\downarrow$}}& COD10K & \textbf{0.017} & 0.026 & 0.008 & 0.017 & 0.017 & 0.000 \\
    & CAMotion & 0.033 & \textbf{0.031} & 0.006 & 0.031 & 0.020 & 0.011 \\
    & MoCA-Mask & 0.040 & 0.044 & \textbf{0.009} & 0.009 & 0.042 & 0.033 \\
    \cmidrule(lr){2-8}
    & Mean others & 0.037 & 0.035 & 0.007 & 0.019 & 0.026 & 0.007 \\
    \bottomrule          
    \end{tabular}
    \label{tab:table4}
\end{table*}


\subsection{Benchmarks}

\textbf{Baseline.} We select 20 cutting-edge baselines, including \textbf{(i)} 14 COD methods, \textit{i.e.}, MGL-R~\cite{zhai2021mutual}, PFNet~\cite{mei2021camouflaged}, UGTR~\cite{yang2021uncertainty}, SegMaR~\cite{jia2022segment}, ZoomNet~\cite{pang2022zoom}, SINet-v2~\cite{fan2021concealed}, FSPNet~\cite{huang2023feature}, PUENet~\cite{10159663}, PopNet~\cite{wu2023source}, DaCOD~\cite{DBLP:conf/mm/WangYYWCZ23}, HGINet~\cite{DBLP:journals/tip/YaoSXWC24}, CamoDiffusion~\cite{DBLP:journals/pami/SunCLSLJ25}, RUN~\cite{DBLP:conf/icml/HeZX0T0KF0F25} and ESCNet~\cite{Ye_2025_ICCV} \textbf{(ii)} six VCOD methods, \textit{i.e.}, SAM2~\cite{DBLP:conf/iclr/RaviGHHR0KRRGMP25}, SLT-Net~\cite{DBLP:conf/cvpr/ChengXFZHDG22}, SAM-PM~\cite{DBLP:conf/cvpr/MeeranTM22}, ZoomNeXt~\cite{DBLP:journals/pami/PangZXZL24}, EMIP~\cite{DBLP:journals/tip/ZhangXJWFZ25}, and CamSAM2~\cite{zhou2025camsam}.

\textbf{Quantitative comparison.} We evaluate 19 selected state-of-the-art methods on CAMotion and MoCA-Mask testing datasets, and present the quantitative performance in Table \ref{tab:table3}. 
Due to the variations of network architecture, input resolutions, modalities, as well as pre-processing techniques, we make the best effort to ensure a fair comparison on both datasets. Regarding CAMotion, we surprisingly observe that the image-level COD method HGINet \cite{DBLP:journals/tip/YaoSXWC24} achieves SOTA on most of the metrics, even surpassing video-based methods like ZoomNeXt \cite{DBLP:journals/pami/PangZXZL24}. Specifically, it achieves performance gains of 6.9\%, 2.4\%, 6.1\%, 5.4\% and 5.4\% in terms of $F_{\beta}^{w}$, $E_{\phi}^{m}$, $\mathit{M}$, $\mathrm{mDic}$ and $\mathrm{mIoU}$, respectively, compared to the current state-of-the-art VCOD method ZoomNeXt. However, ZoomNeXt achieves better performance against HGINet on MoCA-Mask and demonstrates more balanced performance across multiple datasets, which suggests that ZoomNeXt can leverage temporal cues more effectively while still exhibiting limited capability in camouflaged object discrimination. 


\begin{figure*}[t]
\centering
\includegraphics[width=\textwidth]{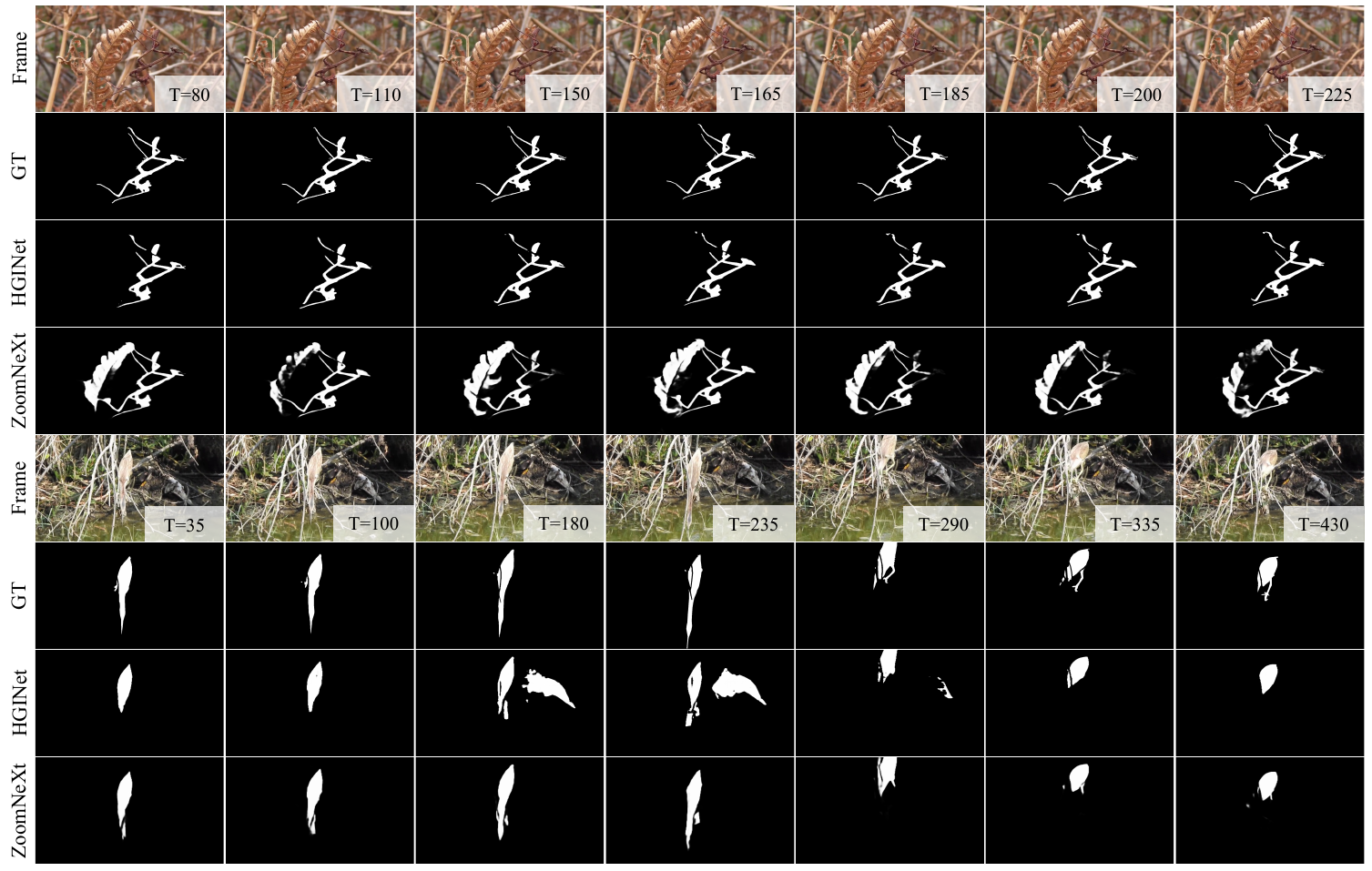}
\caption{Visual comparison with state-of-the-art methods in challenging scenarios, \textit{i.e.}, shape complexity (Rows 1-4) and occlusion (Rows 5-8). Please zoom in for details.}
\label{fig:qualitative}
\end{figure*}


\begin{figure*}[t]
\centering
\includegraphics[width=6in]{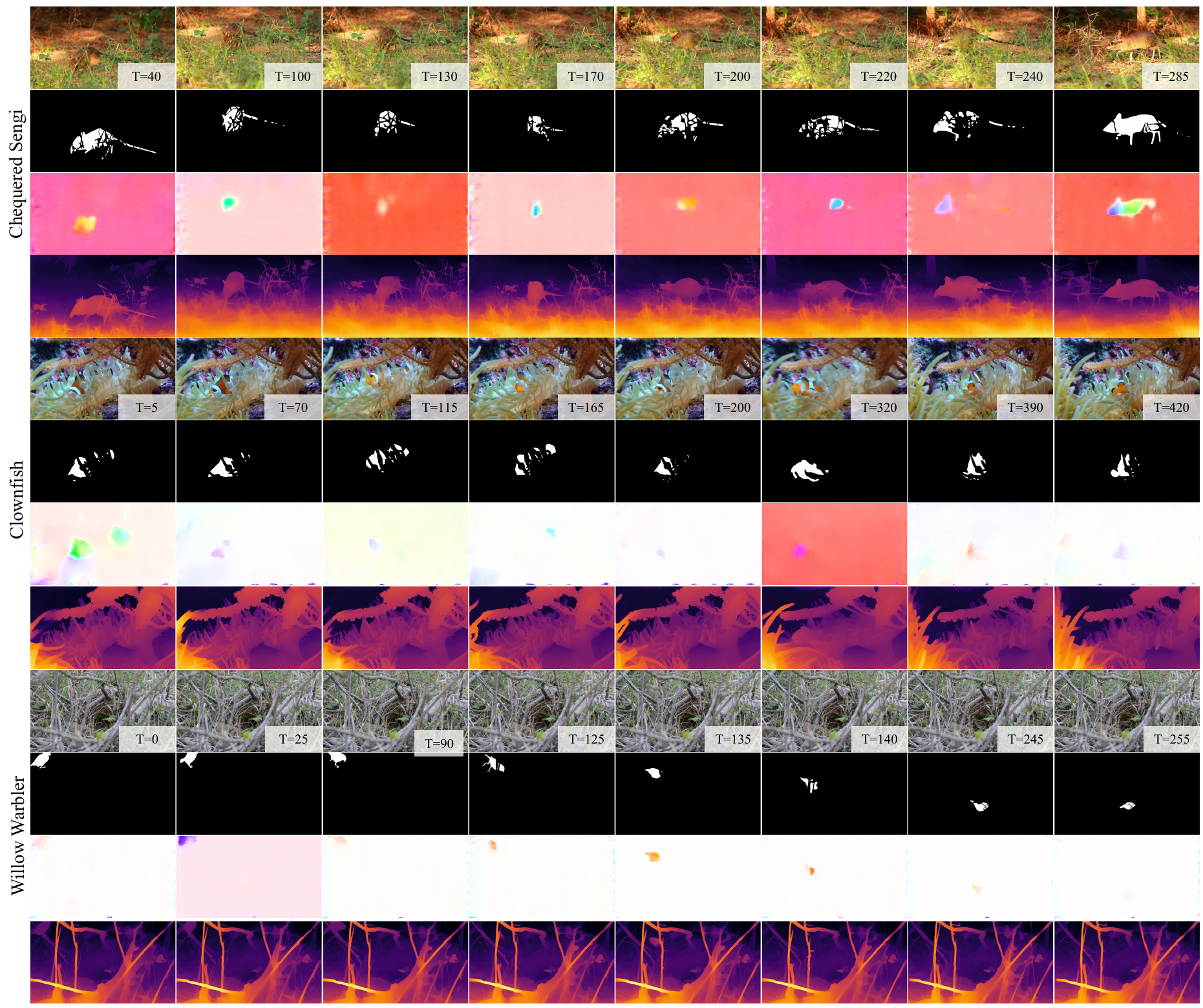}
\caption{Optical flow and depth properties visualization. Each group comprises the original image, pixel-level annotation, optical flow, and depth map. Please zoom in for details.}
\label{fig:depth}
\end{figure*}


Additionally, owing to the diversity of object scales in our dataset, the evaluation results of the SOTA methods on CAMotion are more stable, especially for the $\mathit{M}$ metric relative to other evaluation indicators. In contrast, MoCA-Mask tends to exhibit extremely low $\mathit{M}$ while significantly worse performance on the remaining five metrics. This imbalance can be attributed to the fact that the MoCA-Mask test set consists almost exclusively of small objects and lacks scale diversity, which consequently leads to heavily biased evaluation results. Moreover, the superior performance of HGINet on CAMotion further highlights a critical limitation of existing VCOD methods: their inability to simultaneously preserve accurate camouflaged object detection and reliable temporal consistency. Moreover, the significant performance gap between existing image COD datasets and CAMotion, along with the ineffectiveness of SAM2~\cite{DBLP:conf/iclr/RaviGHHR0KRRGMP25} and CamSAM2~\cite{zhou2025camsam}, highlights the difficulties of detecting camouflaged objects in consecutive video sequences. We believe that CAMotion opens up a broad and meaningful research space, and we strongly encourage the community to conduct further research in these underexplored areas.

\textbf{Qualitative comparison.}
As shown in Fig. \ref{fig:qualitative}, we perform the visual comparison of HGINet \cite{DBLP:journals/tip/YaoSXWC24} and ZoomNeXt \cite{DBLP:journals/pami/PangZXZL24} in two typical scenarios, shape complexity (Rows 1-4) and occlusion (Rows 5-8). Overall, both methods can identify the location and shapes of camouflaged objects in a subset of specific video frames. However, they still suffer from the presence of highly confusing and distracting surrounding backgrounds, which degrade the segmentation performance. As shown in Rows 1-4, HGINet possesses superior discriminative ability in locating and segmenting camouflaged objects from distracting backgrounds against ZoomNeXt. In contrast, ZoomNeXt tends to propagate distracting context across subsequent frames because of its limited discriminative ability. However, HGINet fails to maintain consistent object localization, even though the camouflaged object is well-identified in previous frames (see Row 7). In contrast, Row 8 demonstrates the superior results obtained by ZoomNeXt, as it leverages temporal information to enhance temporal consistency. Such analyses reveal that current methods struggle to balance the discriminative capability and temporal consistency.

\textbf{Optical flow and depth properties.}
We employ GMFlow~\cite{DBLP:journals/pami/XuZCRYTG23} and Depth Anything V2~\cite{DBLP:conf/nips/YangKH0XFZ24} to estimate optical flow and depth map, respectively, with the results visualized in Fig.~\ref{fig:depth}. In the cases of \emph{chequered sengi}, \emph{clownfish} and \emph{willow warbler}, we observe that the optical flow provides informative partial camouflage cues in moving object scenarios, while the depth map can also reveal camouflaged object contours to some extent. However, in scenes with camera pose variation, limited object motion, and low depth contrast between camouflaged objects and surrounding background, the estimated optical flow and depth map fail to provide effective guidance for camouflaged object detection. Additional examples of the optical flow and depth maps are provided in Appendix B.1.


\begin{figure*}[t]
    \centering
    \begin{minipage}[t]{0.469\linewidth}
        \centering
        \includegraphics[width=\linewidth]{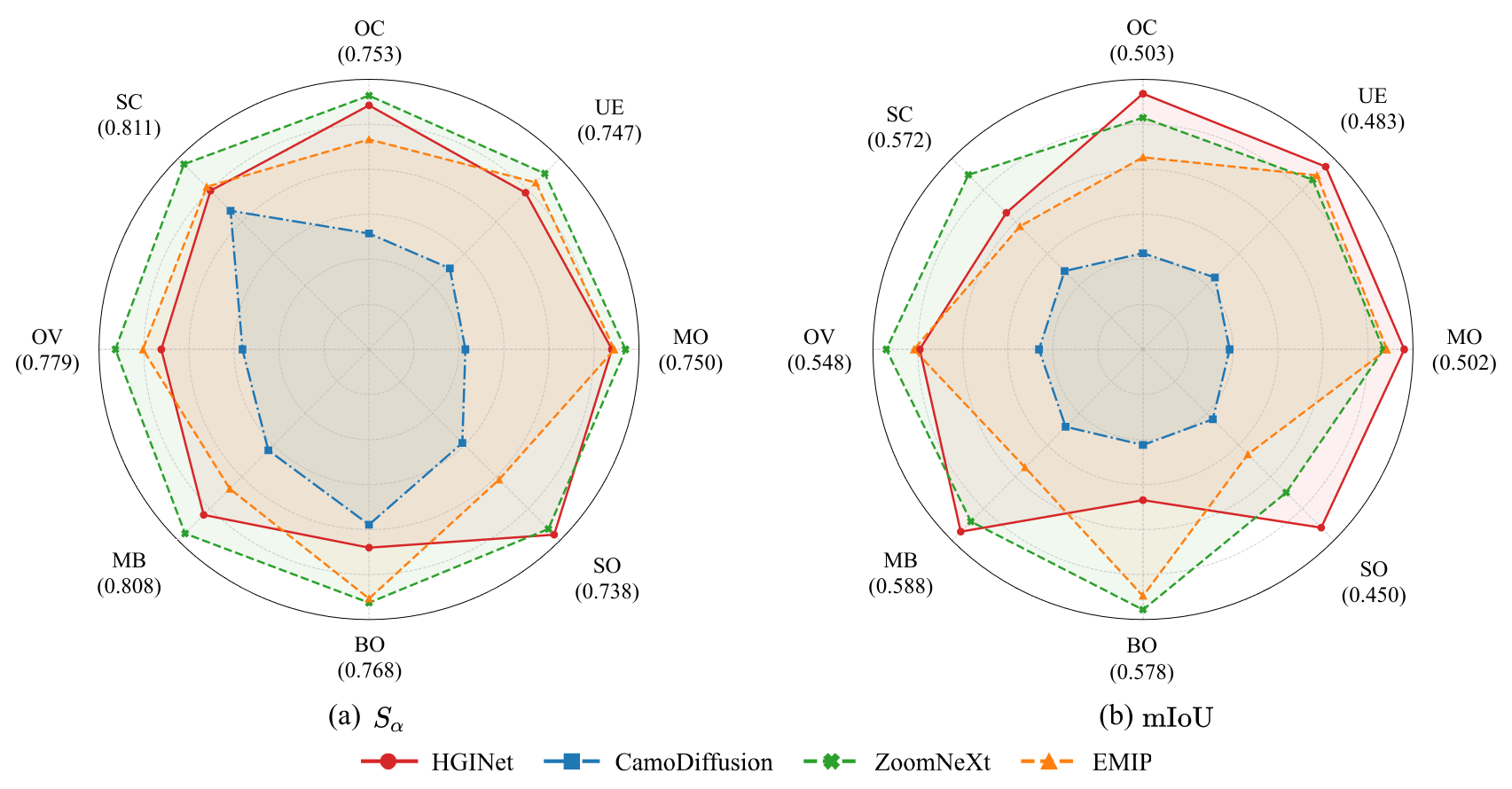}
        \caption{Visualization of SOTA method performances on different challenging attributes under (a) $S_{\alpha}$ and (b) $\mathrm{mIoU}$.}
        \label{fig:attribute}
    \end{minipage}
    \hfill
    \begin{minipage}[t]{0.52\linewidth}
        \centering
        \includegraphics[width=\linewidth]{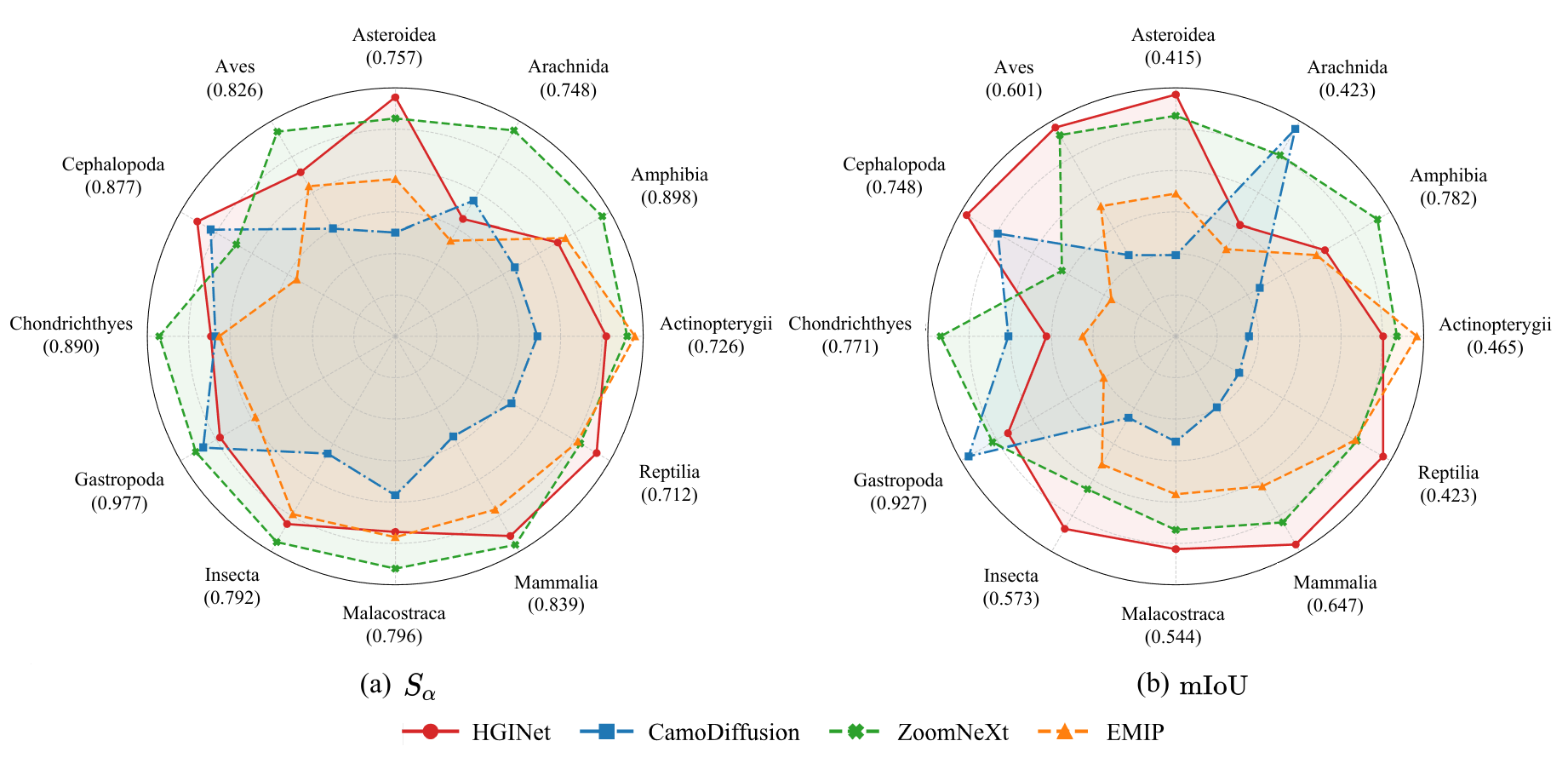}
        \caption{Visualization of SOTA method performances on different classes in terms of (a) $S_{\alpha}$ and (b) $\mathrm{mIoU}$.}
        \label{fig:category}
    \end{minipage}
\end{figure*}

\begin{figure*}[t]
    \centering
    \begin{minipage}[t]{0.36\linewidth}
        \centering
        \includegraphics[width=\linewidth]{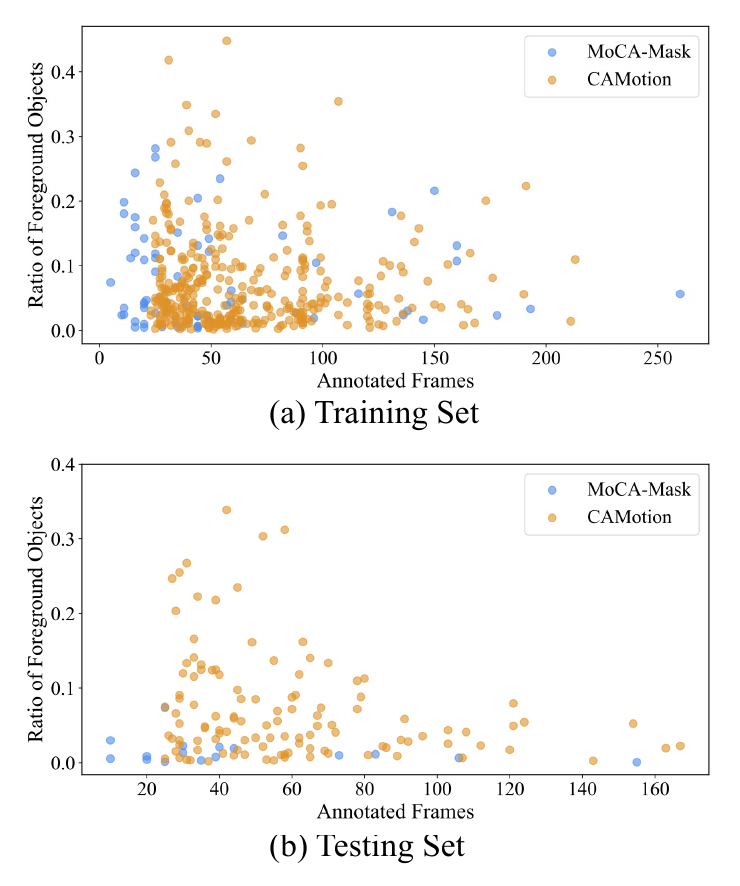}
        \caption{Scale distribution comparison of CAMotion and MoCA-Mask.}
        \label{fig:image1}
    \end{minipage}
    \hfill
    \begin{minipage}[t]{0.63\linewidth}
        \centering
        \includegraphics[width=\linewidth]{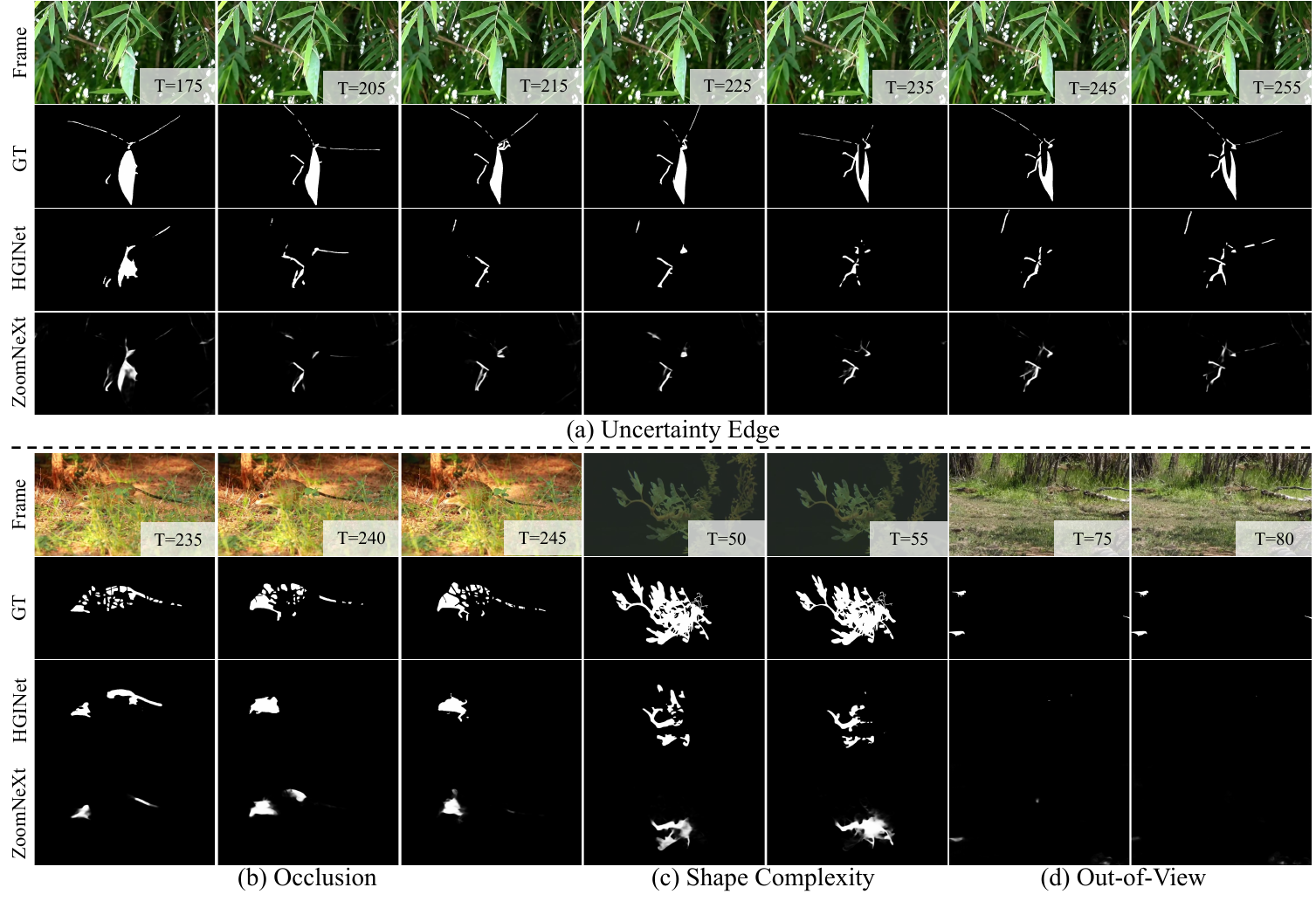}
        \caption{Failure cases on both HGINet and ZoomNeXt in several challenging scenarios. Please zoom in for details.}
        \label{fig:image2}
    \end{minipage}
\end{figure*}




\subsection{Dataset Analysis}

\textbf{Cross-dataset generalization.}
Since the generalization ability and difficulty of a dataset play significant roles in both training and evaluation, we investigate these two aspects on COD10K, our CAMotion and MoCA-Mask datasets, using the cross-dataset analysis method~\cite{DBLP:conf/cvpr/TorralbaE11}, \textit{i.e.}, train a model on one dataset and test it on all selected datasets. For a fair comparison, we use the image version of the recently proposed ZoomNeXt as the base model and reorganize both CAMotion and MoCA-Mask into image-level datasets, so that all datasets can be evaluated under the same training and evaluation settings. ZoomNeXt is then trained on each dataset until the loss becomes stable. 

Table~\ref{tab:table4} shows the results of $S_{\alpha}$ and $\mathit{M}$. Each row presents performance trained on a specific dataset and evaluated on all selected datasets, \textit{i.e.}, COD10K, CAMotion and MoCA-Mask, reflecting the generalization capability of the training dataset. Each column shows the performance of ZoomNeXt tested on a particular dataset, highlighting the difficulty of each dataset. As expected, CAMotion exhibits greater difficulty while providing stronger generalization capability, particularly when evaluated against the large-scale COD benchmark COD10K, under both $S_{\alpha}$ and $\mathit{M}$. Take $\mathit{M}$ as an example, CAMotion is the only dataset where the ``Mean Others'' performance exceeds ``Self" with a 0.011 ``Performance Gap'', indicating a stronger generalization capability on CAMotion. In addition, the ``Mean Others'' $S_{\alpha}$ score on CAMotion is 0.778 lower than 0.809 on COD10K, further confirming the increased difficulty of CAMotion. Moreover, the model trained on CAMotion outperforms the others on the MoCA-Mask testing set in terms of both metrics, demonstrating the generalization ability and diversity of our CAMotion. We also observe that the models trained on COD10K and CAMotion exhibit a better ``Self'' performance versus ``Mean Others". This is because MoCA-Mask has a relatively homogeneous data distribution, as most of the camouflaged objects in the test set are very small, making it less capable of providing a comprehensive performance evaluation. The inconsistency between the low $S_{\alpha}$ and superior $\mathit{M}$ on MoCA-Mask supports our analysis.

\textbf{Attribute-based performances.} To investigate how varying challenging scenes affect the results, we visualize the performance of HGINet \cite{DBLP:journals/tip/YaoSXWC24}, CamoDiffusion \cite{DBLP:journals/pami/SunCLSLJ25}, ZoomNeXt \cite{DBLP:journals/pami/PangZXZL24} and EMIP \cite{DBLP:journals/tip/ZhangXJWFZ25} in eight challenging attributes in terms of $S_{\alpha}$ and $\mathrm{mIoU}$, see Fig. \ref{fig:attribute}. Notably, we observe that the sequences involving small object (SO), uncertainty edge (UE), occlusion (OC) and multiple objects (MO) are significantly more difficult. In contrast, sequences characterized by shape complexity and motion blur tend to yield relatively better performance. Details of other metrics can be found in Appendix C.4.

\textbf{Class-based performances.}
In Fig. \ref{fig:category}, we further visualize the performance of four SOTA methods HGINet \cite{DBLP:journals/tip/YaoSXWC24}, CamoDiffusion \cite{DBLP:journals/pami/SunCLSLJ25}, ZoomNeXt \cite{DBLP:journals/pami/PangZXZL24} and EMIP \cite{DBLP:journals/tip/ZhangXJWFZ25} across different biological camouflaged object classes in terms of $S_{\alpha}$ and $\mathrm{mIoU}$. Overall, the methods exhibit relatively better performance on classes such as \textit{Amphibia}, \textit{Cephalopoda}, \textit{Chondrichthyes} and \textit{Gastropoda}. This is because these classes are more visually distinguishable and exhibit more perceptible texture cues. In contrast, all models perform worse on classes such as \textit{Actinopterygii}, \textit{Asteroidea} and \textit{Reptilia}, where the high visual similarity poses greater challenges for accurate detection. Notably, CamoDiffusion and EMIP exhibit large performance fluctuations across different classes, indicating a weaker generalization capability. In contrast, the other two models demonstrate more consistent trends across all metrics within the 12 classes. Details of other metrics can be found in Appendix C.4.



\begin{table*}[t]
  \footnotesize
  \centering
  \caption{Comparison of HGINet, ZoomNeXt, and EMIP under training annotation intervals of 5 and 10 frames on CAMotion.}
  \label{tab:interval_analysis}
  \begin{tabular}{lcc@{\hspace{15pt}}cccccc}
    \toprule
    \multirow[c]{2}{*}[-0.8ex]{\textbf{Method}} &
    \multirow[c]{2}{*}[-0.8ex]{
        \makecell[c]{\textbf{Training}\\\textbf{Interval}}} &
    \multirow[c]{2}{*}[-0.8ex]{%
      \makecell[c]{\textbf{\#Annotated}\\\textbf{Frames}}} &
    \multicolumn{6}{c}{\textbf{CAMotion (6,775)}} \\
    \cmidrule(lr){4-9}
    & & &
    $S_{\alpha}\uparrow$ &
    $F_{\beta}^{w}\uparrow$ &
    $E_{\phi}^{m}\uparrow$ &
    $\mathit{M}\downarrow$ &
    $\mathrm{mDic}\uparrow$ &
    $\mathrm{mIoU}\uparrow$ \\
    \midrule
    \multirow{2}{*}{HGINet~\cite{DBLP:journals/tip/YaoSXWC24}}
      & 5  & 23,253 & 0.774 & 0.634 & 0.852 & 0.031 & 0.660 & 0.551 \\
      & 10 & 11,715 & 0.773 & 0.623 & 0.875 & 0.033 & 0.666 & 0.547 \\
    \midrule
    \multirow{2}{*}{ZoomNeXt~\cite{DBLP:journals/pami/PangZXZL24}}
      & 5  & 23,253 & 0.779 & 0.593 & 0.832 & 0.033 & 0.626 & 0.523 \\
      & 10 & 11,715 & 0.772 & 0.605 & 0.842 & 0.035 & 0.640 & 0.530 \\
    \midrule
    \multirow{2}{*}{EMIP~\cite{DBLP:journals/tip/ZhangXJWFZ25}}
      & 5  & 23,253 & 0.761 & 0.583 & 0.866 & 0.035 & 0.617 & 0.506 \\
      & 10 & 11,715 & 0.756 & 0.578 & 0.865 & 0.034 & 0.609 & 0.499 \\
    \bottomrule
  \end{tabular}
\end{table*}

\textbf{Ablation on annotation interval.} To further investigate the influence of the training annotation interval, we conduct additional experiments with three representative methods, \emph{i.e.}, image-based HGINet~\cite{DBLP:journals/tip/YaoSXWC24}, video-based ZoomNeXt~\cite{DBLP:journals/pami/PangZXZL24}, and optical-flow-based EMIP~\cite{DBLP:journals/tip/ZhangXJWFZ25}. We increase the training annotation interval from 5 to 10 frames while keeping the other implementation settings unchanged. All variants are evaluated on the same annotated testing frames. As shown in Table~\ref{tab:interval_analysis}, all three methods maintain comparable overall performance when the number of annotated training frames is reduced from 23,253 to 11,715. Specifically, the $S_{\alpha}$ scores of HGINet, ZoomNeXt, and EMIP decrease by 0.001, 0.007, and 0.005, respectively. These results indicate that increasing the training annotation interval from five to ten frames exerts only a marginal impact on the performance of the evaluated methods. This stability occurs because intervals of either five or ten frames provide sufficient temporal correspondence for localizing camouflaged objects in video sequences. Performance will degrade only when the interval is enlarged significantly enough until the temporal correspondences become unreliable.

\textbf{Sequence length analysis.}
To investigate model performance across different sequence lengths, we divide the CAMotion testing set into short ($L\leq200$), medium ($200<L\leq400$), and long ($L>400$) subsets, where $L$ denotes the number of video frames. Detailed statistics are reported in Table~\ref{tab:length_groups}. Then we further evaluate five representative methods on these subsets, including two image-based COD methods, \emph{i.e.}, HGINet~\cite{DBLP:journals/tip/YaoSXWC24} and CamoDiffusion~\cite{DBLP:journals/pami/SunCLSLJ25}, and three video-based methods, \emph{i.e.}, SLT-Net~\cite{DBLP:conf/cvpr/ChengXFZHDG22}, ZoomNeXt~\cite{DBLP:journals/pami/PangZXZL24} and EMIP~\cite{DBLP:journals/tip/ZhangXJWFZ25}. As shown in Table~\ref{tab:length_analysis}, HGINet achieves the best performance across all of the metrics on the long subset, whereas ZoomNeXt obtains the highest $S_{\alpha}$ and $\mathrm{mIoU}$ scores on the short subset. Moreover, all three video-based methods obtain lower $S_{\alpha}$, $F_{\beta}^{w}$, $\mathrm{mDic}$, and $\mathrm{mIoU}$ scores on long sequences than on short and medium sequences. For example, the $\mathrm{mIoU}$ score of ZoomNeXt decreases from 0.591 on short sequences to 0.430 on long sequences. These results show that long sequences remain challenging for current VCOD methods and highlight the value of CAMotion for evaluating model performance across different sequence lengths.

\begin{table}[t]
  \footnotesize
  \centering
  \caption{Length stratification of the CAMotion testing set.
  $L$ denotes the number of frames in each video sequence.}
  \label{tab:length_groups}
  \begin{tabular}{c@{\hspace{15pt}}cccc}
    \toprule
    \makecell{\textbf{Sequence}\\\textbf{Group}} &
    \makecell{\textbf{Video}\\\textbf{Length}} &
    \makecell{\textbf{\#Video}\\\textbf{Sequences}} &
    \makecell{\textbf{\#Video}\\\textbf{Frames}} &
    \textbf{Mean $L$} \\
    \midrule
    Short  & $L\leq200$     & 40  & 6,427  & 160.7 \\
    Medium & $200<L\leq400$ & 54  & 15,509 & 287.2 \\
    Long   & $L>400$        & 21  & 11,734 & 558.8 \\
    \midrule
    Total  & --             & 115 & 33,670 & 292.8 \\
    \bottomrule
  \end{tabular}
\end{table}
\begin{table}[t]
  \setlength{\tabcolsep}{3pt}
  \scriptsize
  \centering
  \caption{Length-stratified performance of five representative methods on the CAMotion testing set.}
  \label{tab:length_analysis}
  \begin{tabular}{lccccccc}
    \toprule
    \multirow[c]{2}{*}[-0.8ex]{\textbf{Method}} &
    \multirow[c]{2}{*}[-0.8ex]{%
      \makecell{\textbf{Sequence}\\\textbf{Length}}} &
    \multicolumn{6}{c}{\textbf{CAMotion Test}} \\
    \cmidrule(lr){3-8}
    & &
    $S_{\alpha}\uparrow$ &
    $F_{\beta}^{w}\uparrow$ &
    $E_{\phi}^{m}\uparrow$ &
    $\mathit{M}\downarrow$ &
    $\mathrm{mDic}\uparrow$ &
    $\mathrm{mIoU}\uparrow$ \\
    \midrule
    \multirow{3}{*}{HGINet~\cite{DBLP:journals/tip/YaoSXWC24}}
      & Short  & 0.764 & 0.642 & 0.826 & 0.044 & 0.656 & 0.556 \\
      & Medium & 0.791 & 0.662 & 0.859 & 0.032 & 0.687 & 0.589 \\
      & Long   & 0.757 & 0.594 & 0.857 & 0.023 & 0.626 & 0.497 \\
    \midrule
    \multirow{3}{*}{CamoDiffusion~\cite{DBLP:journals/pami/SunCLSLJ25}}
      & Short  & 0.713 & 0.532 & 0.740 & 0.050 & 0.548 & 0.475 \\
      & Medium & 0.736 & 0.558 & 0.799 & 0.040 & 0.574 & 0.494 \\
      & Long   & 0.665 & 0.407 & 0.715 & 0.030 & 0.429 & 0.342 \\
    \midrule
    \multirow{3}{*}{SLT-Net~\cite{DBLP:conf/cvpr/ChengXFZHDG22}}
      & Short  & 0.757 & 0.601 & 0.842 & 0.047 & 0.638 & 0.525 \\
      & Medium & 0.767 & 0.592 & 0.860 & 0.041 & 0.631 & 0.525 \\
      & Long   & 0.718 & 0.478 & 0.842 & 0.032 & 0.541 & 0.410 \\
    \midrule
    \multirow{3}{*}{ZoomNeXt~\cite{DBLP:journals/pami/PangZXZL24}}
      & Short  & 0.805 & 0.662 & 0.853 & 0.037 & 0.685 & 0.591 \\
      & Medium & 0.797 & 0.629 & 0.851 & 0.035 & 0.662 & 0.566 \\
      & Long   & 0.740 & 0.507 & 0.796 & 0.028 & 0.545 & 0.430 \\
    \midrule
    \multirow{3}{*}{EMIP~\cite{DBLP:journals/tip/ZhangXJWFZ25}}
      & Short  & 0.776 & 0.646 & 0.864 & 0.040 & 0.668 & 0.560 \\
      & Medium & 0.784 & 0.628 & 0.881 & 0.035 & 0.657 & 0.555 \\
      & Long   & 0.720 & 0.487 & 0.849 & 0.031 & 0.535 & 0.411 \\
    \bottomrule
  \end{tabular}
\end{table}

\textbf{Scale distribution comparison.} To illustrate the rationale for the mismatch between $S_{\alpha}$ and $\mathit{M}$ in Table \ref{tab:table4}, we present the comparison of the scale distribution between CAMotion and MoCA-Mask in the training and testing datasets, see Fig. \ref{fig:image1}. As illustrated in Fig. \ref{fig:image1} (b), the MoCA-Mask testing set only consists of 16 video clips, which are dominated by small objects, the foreground-to-background area ratios for nearly all instances lie within the range of 0 to 0.03. In contrast, our CAMotion testing set contains 115 video clips and exhibits a well-balanced distribution between small and large objects. This discrepancy may partially explain why most models perform poorly in terms of most metrics on the MoCA-Mask testing set but achieve superior performance in terms of $\mathit{M}$. 
By comparison, CAMotion offers a broader and more representative distribution of object scales, making it a more comprehensive and balanced benchmark that reflects real-world scenarios. Fig. \ref{fig:image1} (a) also shows the scale diversity for the CAMotion training set. Notably, MoCA-Mask contains excessive sequences with fewer than 20 annotated frames, which hinders effective model training. From a broader perspective, CAMotion maintains stronger consistency between its training and testing distributions, leading to more reliable and meaningful performance evaluation.

\textbf{Failure cases.} We further present representative failure cases on both HGINet and ZoomNeXt in several challenging scenarios. As depicted in Fig. \ref{fig:image2}, Row 3 shows that HGINet lacks the guidance of temporal cues to segment camouflaged objects across consecutive video frames. Row 4 indicates ZoomNeXt lacks sufficient discriminative ability to break the camouflage and therefore passes the distractive cues to the subsequent frames. Moreover, Fig. \ref{fig:image2} (b) and (c) illustrate failure cases under occlusion and shape complexity scenarios. Although both methods can partially detect camouflaged objects, the segmentation results remain fragmented and imprecise due to the highly similar color and texture patterns shared by the objects and their surroundings, which suggests that both methods still lack sufficient semantic understanding of camouflaged objects. Regarding the out-of-view scenario, Fig. \ref{fig:image2} (d) shows that current models still struggle to perceive fast-moving objects and objects that move out of view. All of these results demonstrate the diversity and difficulty of our proposed CAMotion dataset, emphasizing its value as a benchmark for advancing research in video camouflaged object detection.

\begin{figure}[t]
    \centering
    \includegraphics[width=\linewidth]{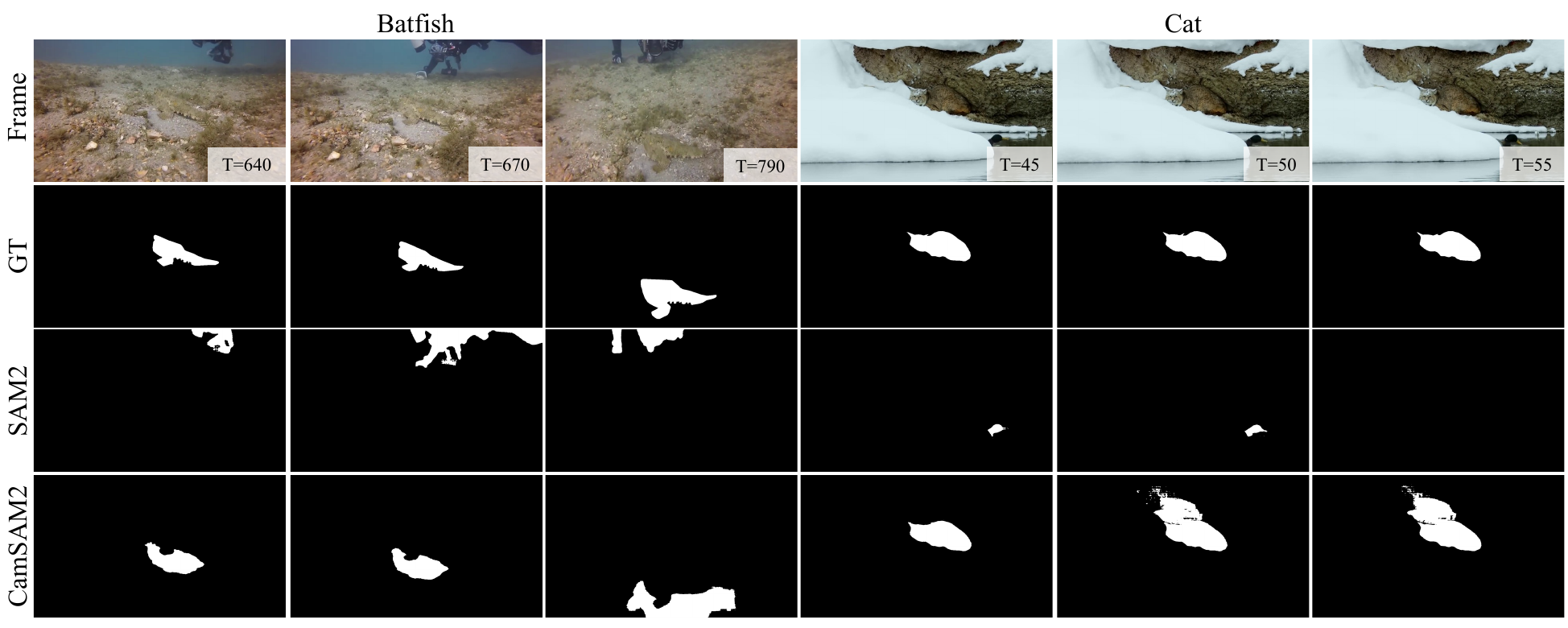}
    \caption{Failure cases of SAM2 and CamSAM2 on CAMotion without first-frame prompts. Please zoom in for details.}
    \label{fig:sam_failure_analysis}
\end{figure}

To further investigate the limitations of SAM-based methods on CAMotion, we present failure cases of SAM2~\cite{DBLP:conf/iclr/RaviGHHR0KRRGMP25} and CamSAM2~\cite{zhou2025camsam} in Fig.~\ref{fig:sam_failure_analysis}. Both methods are evaluated without first-frame prompts following the VCOD setting~\cite{DBLP:conf/cvpr/ChengXFZHDG22}, consistent with the quantitative comparison in Table~\ref{tab:table3}. As shown in Fig.~\ref{fig:sam_failure_analysis}, SAM2 tends to segment salient objects or classify nearly all pixels as background, failing to identify the camouflaged targets. CamSAM2 can locate camouflaged objects in some frames, but produces incomplete masks or includes background regions, leading to inconsistent segmentation across frames. These observations indicate that camouflaged object discovery and temporal consistency remain challenging for these methods under the VCOD setting.

\subsection{Limitation and Future Work}
\label{sec:limitation}
Despite significant advances in COD and VCOD, our experiments reveal a notable trade‑off between camouflaged object discrimination and temporal consistency. Current static COD models, including HGINet and the image‑based variant of ZoomNeXt, achieve strong spatial discriminability on standard COD benchmarks, enabling accurate identification of subtle textural and chromatic differences. However, when deployed on VCOD datasets, such single‑frame COD methods struggle to produce consistent predictions over consecutive frames, even when camouflaged objects are clearly detected in preceding frames. Conversely, temporal-aware methods such as ZoomNeXt excel at capturing temporal cues, producing temporally coherent masks, and handling occlusions and camera motion more robustly. However, they tend to sacrifice the camouflaged discriminability and fail to detect camouflaged objects in several challenging scenarios. 
As a result, existing models fail to simultaneously maintain strong discriminative capability and temporal consistency. The static COD models ignore temporal cues, whereas VCOD algorithms struggle to discriminate challenging camouflaged objects. Bridging this gap is essential for real‑world applications, where both precise localization and stable tracking are required. 

One potential direction is to further explore prompt learning strategies for SAM-based methods. In the supplementary material, we establish a simple baseline that uses HGINet~\cite{DBLP:journals/tip/YaoSXWC24} to generate a first-frame mask prompt for CamSAM2~\cite{zhou2025camsam}, which obtains competitive performance on CAMotion benchmark. Beyond this simple combination, jointly optimizing prompt generation and video segmentation within a unified end-to-end framework is a promising direction for improving camouflaged object discrimination and temporal consistency.

Another direction is to develop dedicated architectures that jointly exploit appearance, optical flow, depth, and other complementary cues. As reported in the supplementary material, the provided depth maps offer useful information for camouflaged object segmentation, while current methods using depth or optical flow still exhibit limited performance. Beyond these cues, long-term trajectories can be explored to model temporal correspondences across frames. Integrating motion cues and trajectory information with appearance features offers a potential direction for improving camouflaged object discrimination and maintaining consistent segmentation across video frames.

In addition, inspired by R2C7K~\cite{DBLP:journals/pami/ZhangYLHFC25}, CAMotion can be extended by providing visual or language prompts to identify and segment corresponding camouflaged objects in videos. However, this extension requires collecting carefully annotated images or language references for each video, which is beyond the scope of the present work. We will explore this direction and develop dedicated datasets and methods for Referring Video Camouflaged Object Detection in the future.

\section{Conclusion}
In this paper, we construct CAMotion, a high-quality benchmark covers a wide range of species for camouflaged motion object detection in the wild. CAMotion comprises various sequences with multiple challenging attributes such as uncertain edge, occlusion, motion blur, and shape complexity, etc. Then we present annotation details and statistical distributions of the dataset, allowing CAMotion to analyze motion characteristics of camouflaged objects across diverse challenging scenarios. Finally, we conduct a comprehensive evaluation of existing SOTA models on the CAMotion dataset and investigate the major challenges in the VCOD task.

\appendix

\section{Datasets/Benchmarks}
\subsection{COD Benchmarks}

\textbf{CHAMELEON.} CHAMELEON is the smallest COD dataset consisting of only 76 images. The images are collected from the Internet using the keyword “camouflaged animal” via Google search. Each image is manually annotated with pixel-level ground truth. Due to its limited size and the absence of formal peer review, CHAMELEON is typically utilized for preliminary validation or as a supplementary benchmark rather than for model training.

\textbf{CAMO.} CAMO is the first officially released COD dataset, containing a total of 2,500 images. It is composed of two equal subsets: the CAMO subset, which includes 1,250 images featuring at least one camouflaged object, and the MS-COCO subset, which includes 1,250 non-camouflaged images used as negative samples. The dataset spans eight super-categories to ensure category diversity. A standard 80\%-20\% split is applied for training and testing.


\begin{figure*}[!t]
\centering
\includegraphics[width=4.5in]{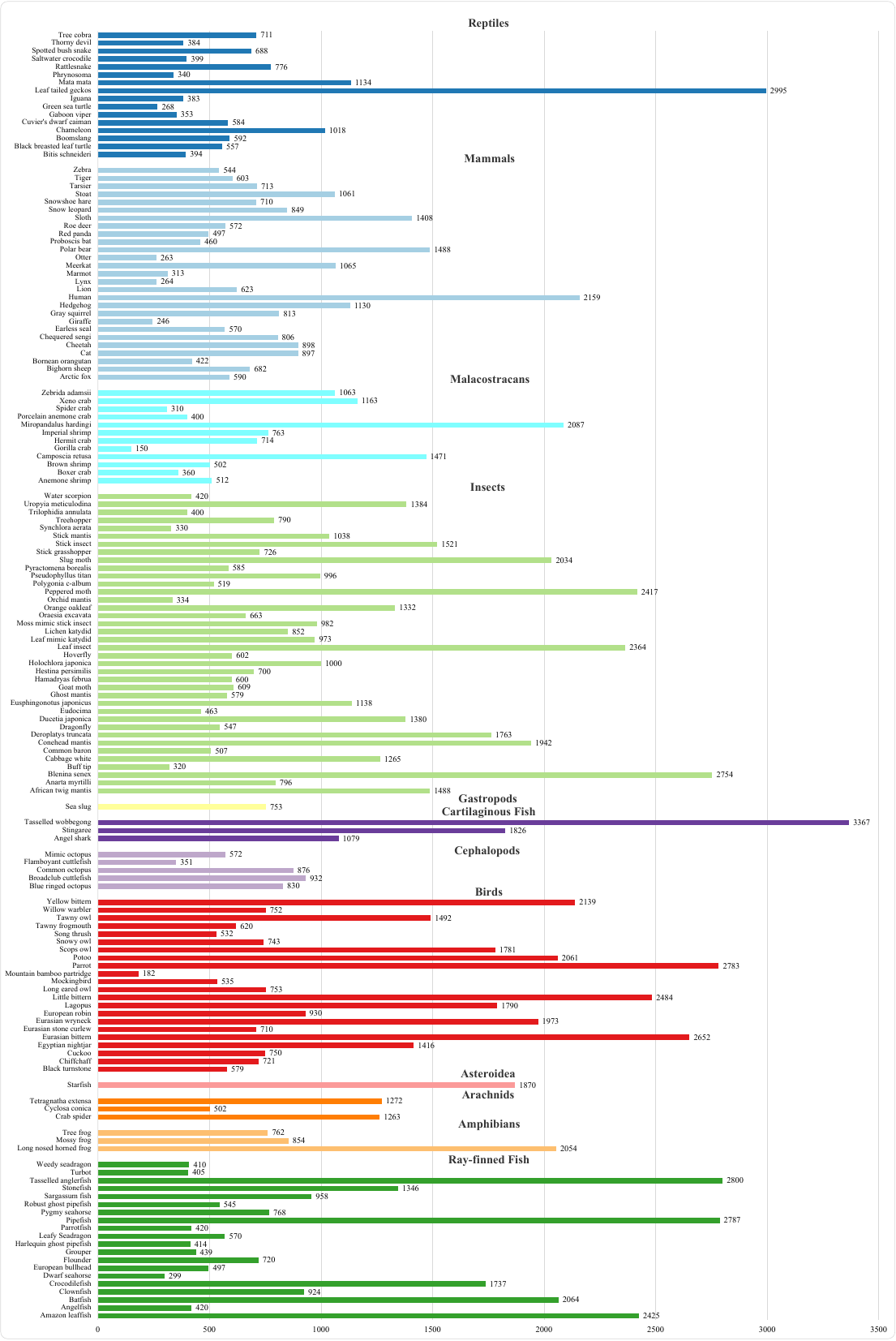}
\caption{Detailed classification of the total frame count across different species. Please zoom in for details.}
\label{fig:counts}
\end{figure*}


\textbf{COD10K.} COD10K consists of 10,000 high-resolution images sourced from various photography platforms, encompassing 10 super-categories and 78 sub-categories. The dataset is divided into three types: 5,066 images containing camouflaged objects, 3,000 background-only images, and 1,934 images featuring non-camouflaged objects. All camouflaged instances are annotated in a hierarchical manner, including category labels, bounding boxes, object-level masks, and instance-level segmentation, thus enabling a wide range of downstream tasks. In most research settings, only the 5,066 images with camouflaged objects are utilized for training and evaluation of COD models.

\textbf{NC4K.} The NC4K dataset contains 4,121 images collected from online resources, all of which are annotated with both object-level and instance-level ground truth masks. Unlike other datasets primarily used for training, NC4K is typically employed as a dedicated test set to assess the generalization performance of COD models. As the largest test-only dataset currently available in the COD field, it provides a comprehensive and diverse benchmark for evaluating the robustness of COD algorithms in real-world scenarios.

\textbf{CAMO++.} CAMO++ is an extended version of the CAMO dataset, expanding both scale and diversity of the dataset. It comprises 5,500 images featuring humans and over 90 distinct animal species, including 2,700 camouflaged images and 2,800 non-camouflaged counterparts. Each image is annotated with a hierarchical labeling scheme that includes meta-category and fine-grained category labels, bounding boxes, and instance-level segmentation masks. Additionally, all instances are annotated with pixel-level ground truth masks through manual labeling. CAMO++ serves as a valuable benchmark not only for camouflaged instance segmentation but also for broader tasks such as semantic camouflage segmentation and video-based camouflaged object detection.


\begin{figure*}[!t]
\centering
\includegraphics[width=5in]{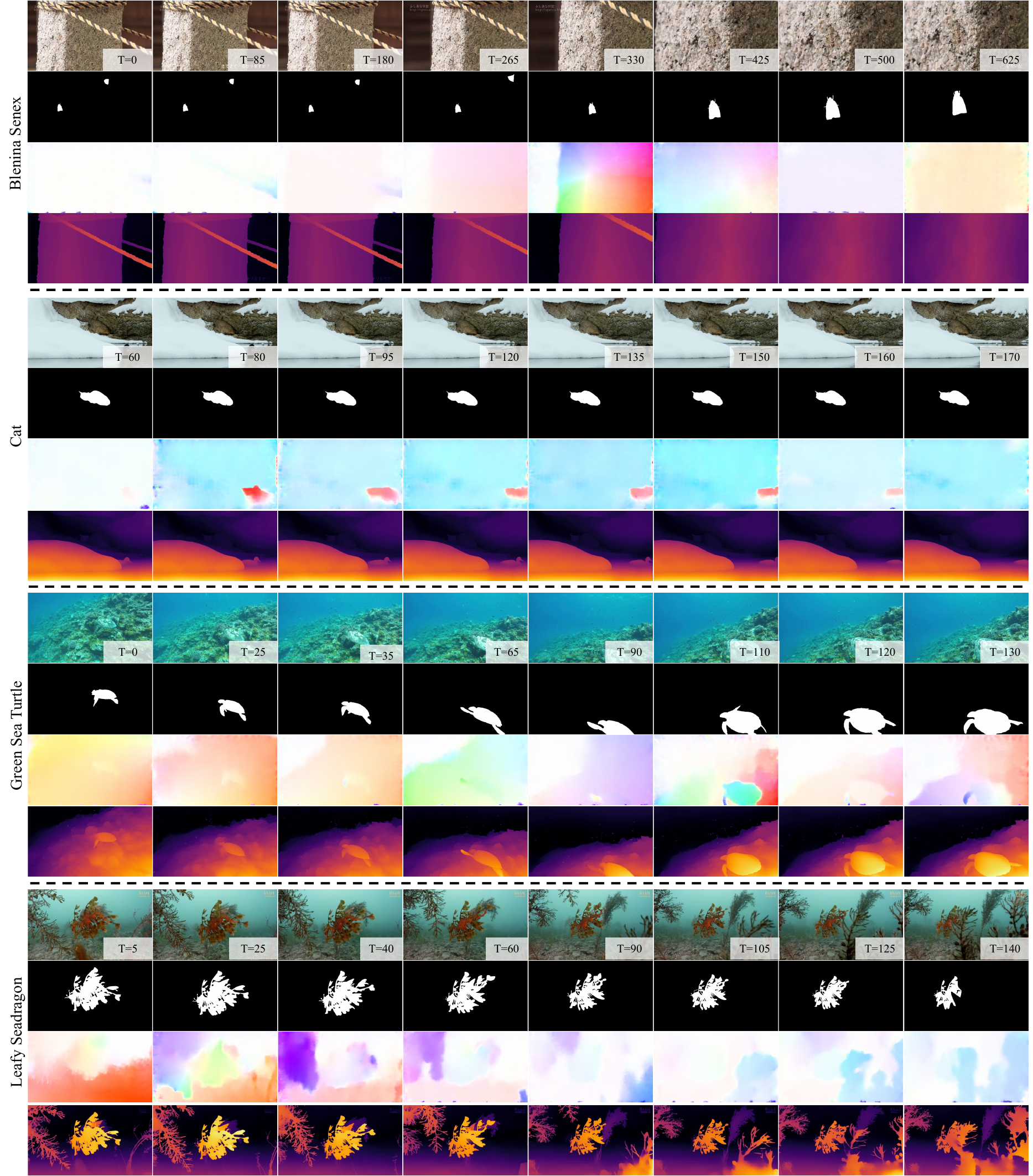}
\caption{Examples of our CAMotion dataset with corresponding pixel-level annotations, optical flow and depth map. Each group is arranged as the original image, optical flow, depth map and pixel-level annotation. Please zoom in for details.}
\label{fig:optical_flow}
\end{figure*}

\subsection{VCOD Benchmarks}

\textbf{CAD.} The Camouflaged Animal Dataset (CAD) is a small-scale dataset specifically designed for camouflaged motion object segmentation. It consists of nine short video sequences collected from publicly available YouTube content. To support temporal analysis, manual pixel-level annotations are provided for every fifth frame within each sequence. Despite its limited size, CAD offers valuable benchmarks for evaluating the temporal consistency and robustness of COD models in dynamic, real-world environments.

\textbf{MoCA‑Mask.}  The original Moving Camouflaged Animals (MoCA) dataset comprises approximately 37,000 frames extracted from 141 YouTube video sequences, predominantly recorded at a resolution of 720 $\times$ 1280 pixels and a frame rate of 24 fps. This dataset includes 67 animal species captured in natural environments, though not all of these animals are camouflaged. The original annotations provide only bounding box labels rather than pixel-level segmentation masks, limiting its effectiveness for evaluating VCOD segmentation tasks. To address this limitation, MoCA was reprocessed into the MoCA-Mask dataset, which contains 87 high-quality video sequences totaling 22,939 frames with dense, pixel-wise annotations on every fifth frame.

\textbf{CamoVid60K}. CamoVid60K~\cite{DBLP:journals/ijcv/VuZSGTY26} contains 218 videos with 62,774 frames covering 70 species, whereas CAMotion contains 474 videos with 149,319 frames covering 151 species. CamoVid60K incorporates CAD~\cite{DBLP:conf/eccv/BideauL16}, MoCA~\cite{DBLP:conf/accv/LamdouarYXZ20}, MoCA-Mask~\cite{DBLP:conf/cvpr/ChengXFZHDG22}, and MVK~\cite{DBLP:conf/mmm/TruongVHLTJY23} together with additional Internet videos, while all videos in CAMotion are collected directly from the Internet. 

With respect to annotation quality, CamoVid60K employs a semi-automatic pipeline: SAM~\cite{DBLP:conf/iccv/KirillovMRMRGXW23} for mask initialization and XMem~\cite{DBLP:conf/eccv/ChengS22} for temporal propagation, followed by manual refinement. This semi-automatic workflow cannot fully guarantee precise object boundaries or temporally consistent ground-truth labels. For CAMotion, we adopt a full manual annotation protocol with multi-stage quality control. Object boundaries are carefully refined through cross-validation and expert review. To enforce temporal consistency, senior annotators further inspect masks overlaid on original video sequences, and annotations with temporal flickering or inconsistent mask transitions are returned for correction. Accordingly, CAMotion outperforms CamoVid60K in both dataset scale and annotation fidelity.

\section{CAMotion Details}

\subsection{Classes and Species}
\label{ref_category}

We present a detailed breakdown of the total frame count according to the taxonomic hierarchy tree, categorized into 12 distinct classes, which are \textit{Ray-finned Fish, Amphibians, Arachnids, Asteroidea, Birds, Cephalopods, Cartilaginous Fish, Gastropods, Insects, Malacostracans, Mammals, and Reptiles} and 151 species (see Fig.~\ref{fig:counts}).

As shown in Fig.~\ref{fig:optical_flow}, we present examples from our CAMotion dataset, showing the original video frames alongside their corresponding optical flow generated using GMFlow~\cite{DBLP:journals/pami/XuZCRYTG23}, depth map derived from Depth Anything V2 \cite{DBLP:conf/nips/YangKH0XFZ24}, and pixel-level annotations. These visualizations collectively demonstrate the richness and reliability of motion cues, depth information, and semantic annotations in our dataset.

\begin{figure*}[!t]
\centering
\includegraphics[width=5in]{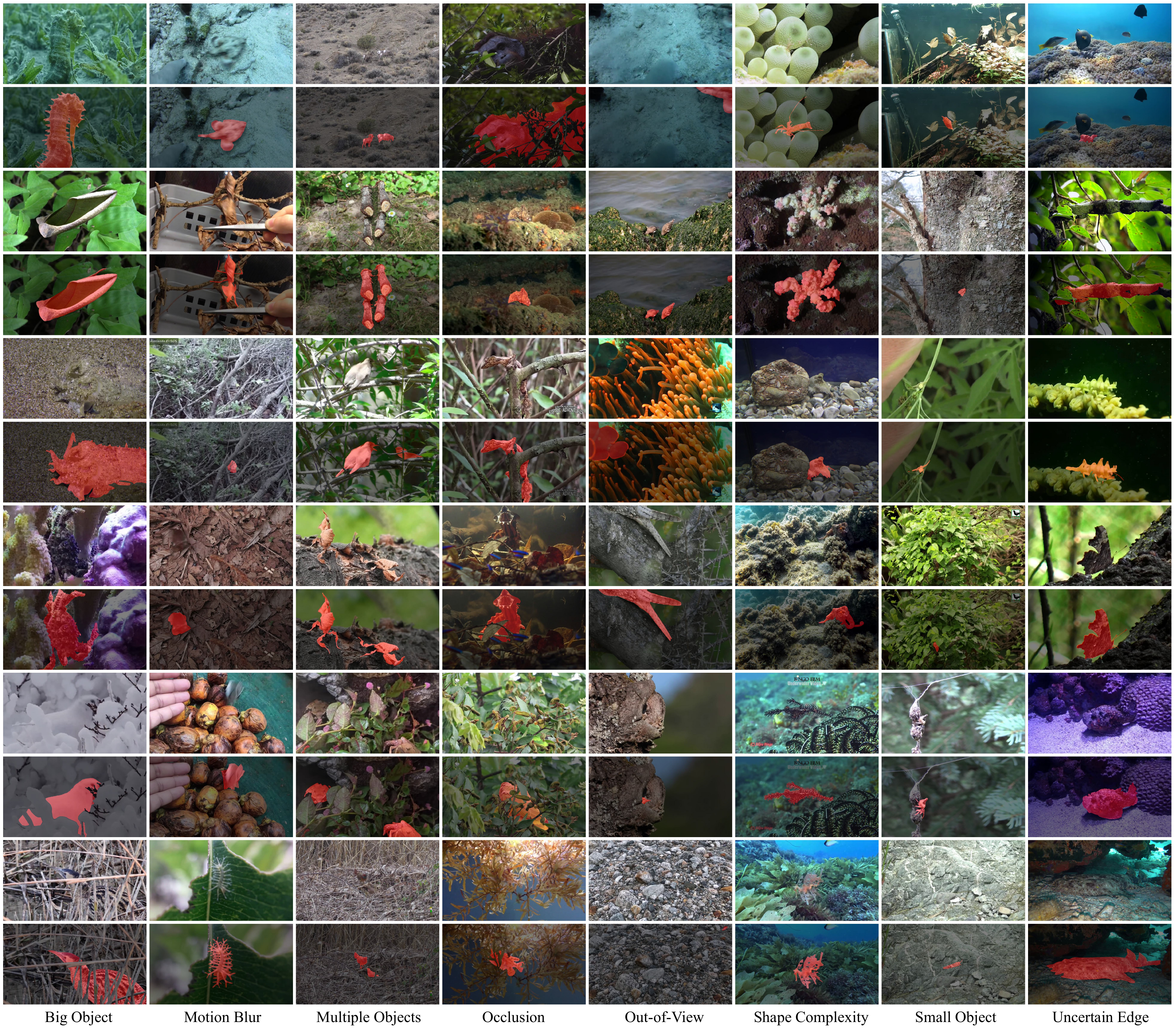}
\caption{Visualization of the challenging attributes in CAMotion. Best viewed in color and zoom in for details.}
\label{fig:attribute_appendix}
\end{figure*}


\begin{figure*}[h]
\centering
\includegraphics[width=5in]{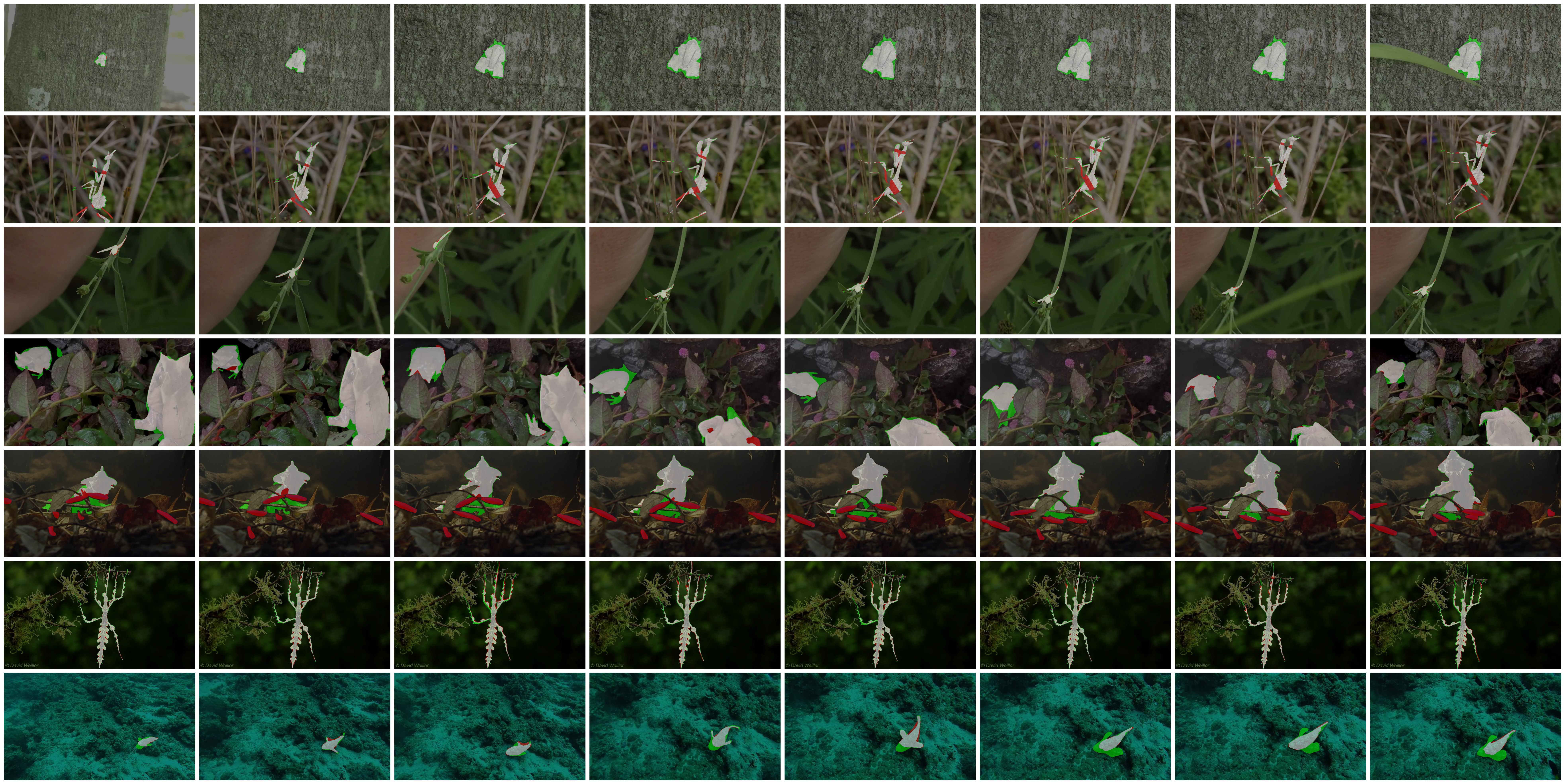}
\caption{Examples of fine-tuning initial annotations. The white color denotes unchanged areas, while the red and green indicate the original and refined annotations, respectively. Please zoom in for details.}
\label{fig:annotation_appendix}
\end{figure*}
\begin{figure*}[h]
\centering
\includegraphics[width=5in]{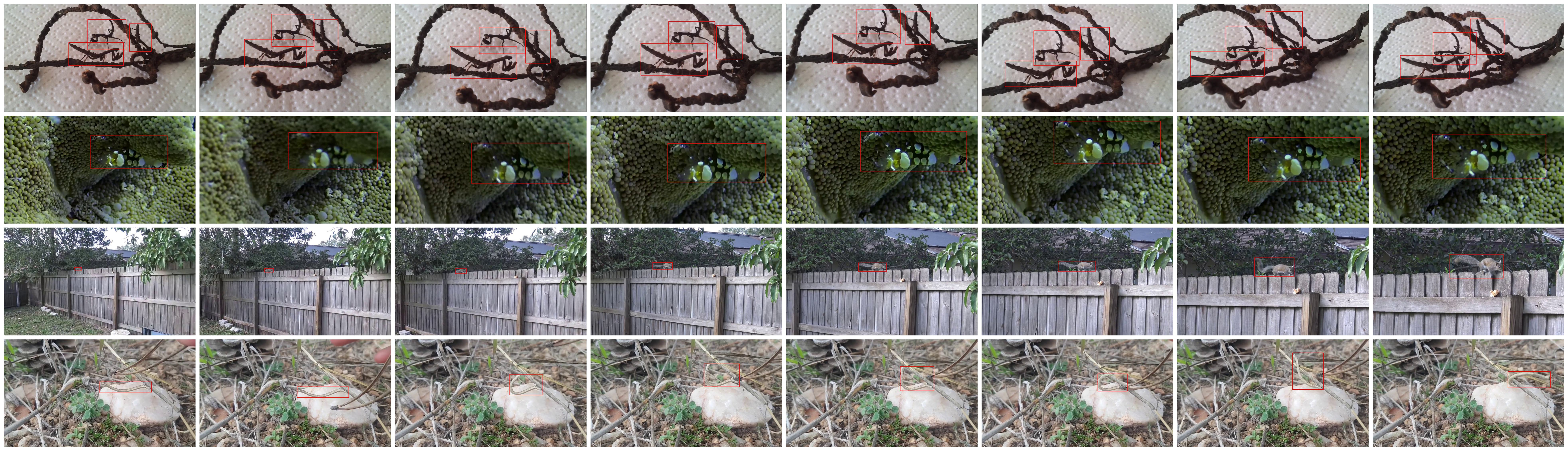}
\caption{Visualization of the bounding box annotations in CAMotion. Best viewed in color and zoom in for details.}
\label{fig:bbox_sequences}
\end{figure*}


\subsection{Attributes}
\label{ref_attr}

As an extension of Fig.~\ref{fig:attribute_show}, Fig.~\ref{fig:attribute_appendix} presents more examples of eight challenging attributes in CAMotion, in terms of \textit{Big Object (BO), Motion Blur (MB), Multiple Objects (MO), Occlusion (OC), Out-of-View (OV), Shape Complexity (SC), Small Object (SO), and Uncertain Edge (UE)}.

\subsection{Annotations}
As shown in Fig.~\ref{fig:annotation_appendix}, we present more challenging frames that are initially labeled inaccurately and corrected during review and validation. These examples demonstrate our efforts to construct a high-quality VCOD dataset with temporally consistent and accurate annotations. In particular, we carefully refine ambiguous object boundaries, especially in regions where camouflaged objects are highly similar to the background, and further correct challenging cases involving occlusion, partially visible objects, etc. To ensure temporal consistency, we overlay the annotated masks onto the original videos and ask three experts to inspect the resulting videos frame by frame. If temporal flickering, inconsistent mask transitions, or inaccurate annotations are observed, the corresponding masks are sent back for further refinement. Through this rigorous annotation and verification procedure, CAMotion provides reliable pixel-level ground truth annotations for training and evaluating VCOD methods. Additionally, we visualize the bounding box annotations of our CAMotion dataset in Fig.~\ref{fig:bbox_sequences}. We also encourage the research community to further explore and customize annotations based on the provided masks and bounding boxes according to specific research objectives.





\section{Experiments}
\label{sec:moreexp}

\subsection{Evaluation Metrics}
To comprehensively evaluate the performance of SOTA methods on the proposed CAMotion, we adopt six widely used quantitative metrics \cite{DBLP:conf/cvpr/ChengXFZHDG22}: structure measure ($S_\alpha$) \cite{Cheng2021sMeasure}, weighted F-measure ($F^w_\beta$) \cite{6909433}, enhanced-alignment measure ($E_{\phi}^{m}$) \cite{ijcai2018p97}, Mean absolute error ($\mathit{M}$), mean dice coefficient ($\mathrm{mDic}$), and mean intersection over union ($\mathrm{mIoU}$). These metrics assess prediction quality from the aspects of structure similarity, pixel-wise accuracy, and spatial overlap.

\textbf{Structure measure ($S_\alpha$).} Considering that camouflaged objects have complex shapes, $S_\alpha$ is used for evaluating structural similarity between prediction and ground truth by combining region-aware ($S_r$) and object-aware ($S_o$) components, which is defined as: 
\begin{equation}
S_\alpha = (1 - \alpha) \cdot S_o + \alpha \cdot S_r, 
\end{equation}
where $\alpha \in [0, 1]$ is the balance parameter and is set to 0.5 in the experiments. Note that the region-aware structural similarity ($S_r$) is designed to assess the object-part structure similarity against the GT masks. It recursively divides each of the predicted and GT masks into four blocks using horizontal and vertical cutoff lines that intersect at the centroid of the GT foreground, and calculates the structural similarity measure (SSIM) of each block using:
\begin{equation}
S_r = \sum_{k=1}^{K} w_k \cdot \mathrm{SSIM}(k),
\end{equation}
where $w_k$ is the assigned weight of each block proportional to the GT foreground region this block covers, and $K$ is the total number of blocks. The object-aware structural similarity ($S_o$) is designed mainly to capture the sharp foreground-background contrast ($S_{FG}$) and uniform distribution ($S_{BG}$) between predicted and GT masks, which is defined as: 
\begin{align}
S_o &= \mu \cdot S_{FG} + (1 - \mu) \cdot S_{BG}, \\
S_{FG} &= \frac{2\overline{x}_{FG}}{(\overline{x}_{FG})^{2}+1+2\lambda \cdot \sigma_{x_{FG}}},   \\
S_{BG} &= \frac{2\overline{x}_{BG}}{(\overline{x}_{BG})^{2}+1+2\lambda \cdot \sigma_{x_{BG}}} ,
\end{align}
where $\sigma_{x}$ and $\overline{x}$ denote the standard deviation and mean of the predicted binary mask, respectively. $\mu$ is the ratio of foreground area in GT to image area, $\lambda$ is a constant to balance the two terms, and $FG$ and $BG$ represent foreground and background area, respectively.

\textbf{Weighted F-measure ($F^w_\beta$)} improves the traditional F-measure by assigning different weights to precision and recall based on spatial importance. It is more sensitive to errors near object boundaries and provides a more reliable segmentation evaluation, which is defined as:
\begin{equation}
F_\beta^w = \frac{(1 + \beta^2) \cdot \text{Precision}^w \cdot \text{Recall}^w}{\beta^2 \cdot \text{Precision}^w + \text{Recall}^w},
\end{equation}
where $\beta^2$ is typically set to 0.3 and thereby placing more emphasis on precision than recall.

\textbf{Enhanced-alignment measure ($E_\phi$)} integrates both pixel-level matching and image-level statistics, simulating human visual perception. It measures alignment quality from both structural and statistical perspectives, especially suitable for binary segmentation tasks, which is defined as: 
\begin{equation}
E_\phi = \frac{1}{W \times H} \sum_{x=1}^{W} \sum_{y=1}^{H} \phi(\frac{2\varphi_G\circ \varphi_P}{\varphi_G\circ \varphi_G+\varphi_P\circ \varphi_P} ),
\end{equation}
where $\phi(\cdot)$ denotes the enhanced alignment matrix, $\circ$ is the Hadamard product. $\varphi$ is the deviation matrix, which represents the distance between each pixel value of the input binary mask and its global mean. $P$ and $G$ denote the predicted binary mask and GT, respectively.

\textbf{Mean absolute error ($\mathit{M}$)} computes the average absolute difference between the predicted map and the ground truth mask at the pixel level. It provides a straightforward measure of prediction error with smaller values indicating better performance, which is defined as: 
\begin{equation}
\mathit{M} = \frac{1}{N} \sum_{i=1}^{N} |P_i - G_i|.
\end{equation}

\textbf{Mean dice coefficient} ($\mathrm{mDic}$) quantifies the spatial overlap between predicted and ground truth masks using the Dice formula. It emphasizes the consistency of foreground extraction and is widely adopted in segmentation evaluation. Dice coefficient is defined as: 
\begin{equation}
\text{Dice} = \frac{2|P \cap G|}{|P| + |G|} = \frac{2TP}{2TP + FP + FN}.
\end{equation}

\textbf{Mean intersection over union} ($\mathrm{mIoU}$) calculates the ratio of the intersection and union of predicted and ground truth regions. As a stricter metric, it is highly representative in segmentation tasks for evaluating spatial accuracy and generalization. IoU is defined as: 
\begin{equation}
\text{IoU} = \frac{TP}{TP + FP + FN}.
\end{equation}

$\mathbf{dtSSD}$ is a temporal consistency metric evaluated in the video benchmark~\cite{DBLP:conf/bmvc/ErofeevGVFW15}, which measures the discrepancy between the temporal derivatives of the predictions and the ground-truth masks:
\begin{equation}
\mathbf{dtSSD}=
\frac{1}{T}\sum_{t=1}^{T}\sqrt{\sum_{p\in\Omega}\left(\frac{\partial P_{p,t}}{\partial t}-\frac{\partial G_{p,t}}{\partial t}\right)^2},
\end{equation}
where $P_{p,t}$ and $G_{p,t}$ denote the prediction and ground-truth value at pixel $p$ in frame $t$, respectively. $\Omega$ denotes the set of all pixels in each frame, and $T$ denotes the number of frames. Lower values indicate better consistency between the temporal changes in the predictions and the ground-truth masks. Since CAMotion provides pixel-level annotations at a 5-frame interval, we compute $\mathbf{dtSSD}$ between the annotated frames within each video.
\subsection{Baseline}

\begin{figure*}[t]
\centering
\includegraphics[width=6.2in]{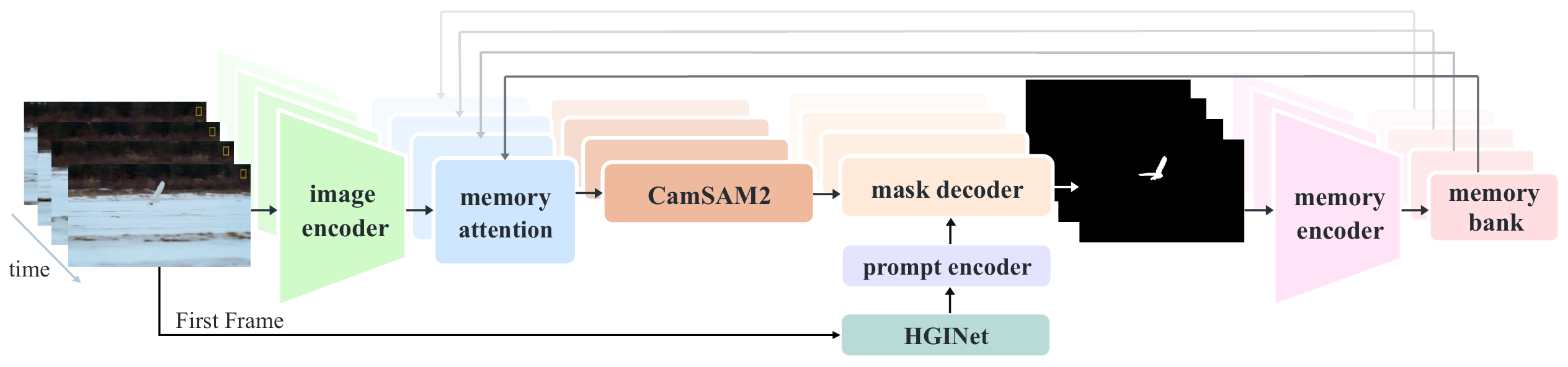}
\caption{Architecture of our simple baseline. For simplicity, the additional components introduced by CamSAM2 are shown in a single block.}
\label{fig:combined_baseline}
\end{figure*}

After constructing the dataset, we also establish a simple baseline, which takes the output of HGINet~\cite{DBLP:journals/tip/YaoSXWC24} as the first-frame prompt for CamSAM2~\cite{zhou2025camsam}, which obtains competitive performance on CAMotion benchmark. As shown in Fig.~\ref{fig:combined_baseline}, HGINet predicts the first-frame mask of each video, which is then fed into CamSAM2 as a prompt for subsequent video segmentation. 

\begin{table}[h]
\scriptsize
\centering
\caption{Quantitative comparison of our simple baseline with representative methods on CAMotion. $^{\ddagger}$ indicates that the first-frame annotation is removed during both training and testing for fair comparison. Our baseline uses first-frame masks predicted by HGINet as prompts.}
\label{tab:combined_baseline}
\setlength{\tabcolsep}{4pt}
\begin{tabular}{lccccccc}
\toprule
\multirow{2}{*}[-0.8ex]{\textbf{Method}} &
\multicolumn{7}{c}{\textbf{CAMotion (6,775)}} \\
\cmidrule(lr){2-8}
& $S_{\alpha}\uparrow$
& $F_{\beta}^{w}\uparrow$
& $E_{\phi}^{m}\uparrow$
& $\mathit{M}\downarrow$
& $\mathrm{mDic}\uparrow$
& $\mathrm{mIoU}\uparrow$
& $\mathbf{dtSSD}\downarrow$ \\
\midrule
HGINet~\cite{DBLP:journals/tip/YaoSXWC24}
& 0.774 & \underline{0.634} & \underline{0.852}
& \textbf{0.031} & \underline{0.660} & \underline{0.551}
& 0.114 \\

ZoomNeXt~\cite{DBLP:journals/pami/PangZXZL24}
& \textbf{0.779} & 0.593 & 0.832
& \underline{0.033} & 0.626 & 0.523
& \textbf{0.097} \\

EMIP~\cite{DBLP:journals/tip/ZhangXJWFZ25}
& 0.761 & 0.583 & \textbf{0.866}
& 0.035 & 0.617 & 0.506
& 0.105 \\

SAM-PM~\cite{DBLP:conf/cvpr/MeeranTM22}$^{\ddagger}$
& 0.741 & 0.550 & 0.802
& 0.047 & 0.589 & 0.488
& 0.101 \\

CamSAM2~\cite{zhou2025camsam}$^{\ddagger}$
& 0.626 & 0.378 & 0.701
& 0.075 & 0.393 & 0.328
& 0.116 \\
\midrule
Ours
& \underline{0.775} & \textbf{0.646} & 0.833
& 0.035 & \textbf{0.669} & \textbf{0.570}
& \underline{0.098} \\
\bottomrule
\end{tabular}
\end{table}

As shown in Table~\ref{tab:combined_baseline}, our proposed method achieves the highest $F_{\beta}^{w}$, $\mathrm{mDic}$, and $\mathrm{mIoU}$ scores among the compared methods. Specifically, it achieves the $\mathrm{mIoU}$ score of 0.570, compared with 0.551, 0.523, and 0.506 for HGINet, ZoomNeXt, and EMIP, respectively. These results indicate that even pseudo visual prompts, such as masks or points, can provide beneficial guidance for camouflaged object discovery across subsequent video frames. Beyond this prompt-based baseline, jointly exploiting appearance, optical flow, and depth remains a direction for future research, as discussed in the ``Limitations and Future Work'' section of the main paper.


\begin{figure*}[!t]
\centering
\includegraphics[width=6.2in]{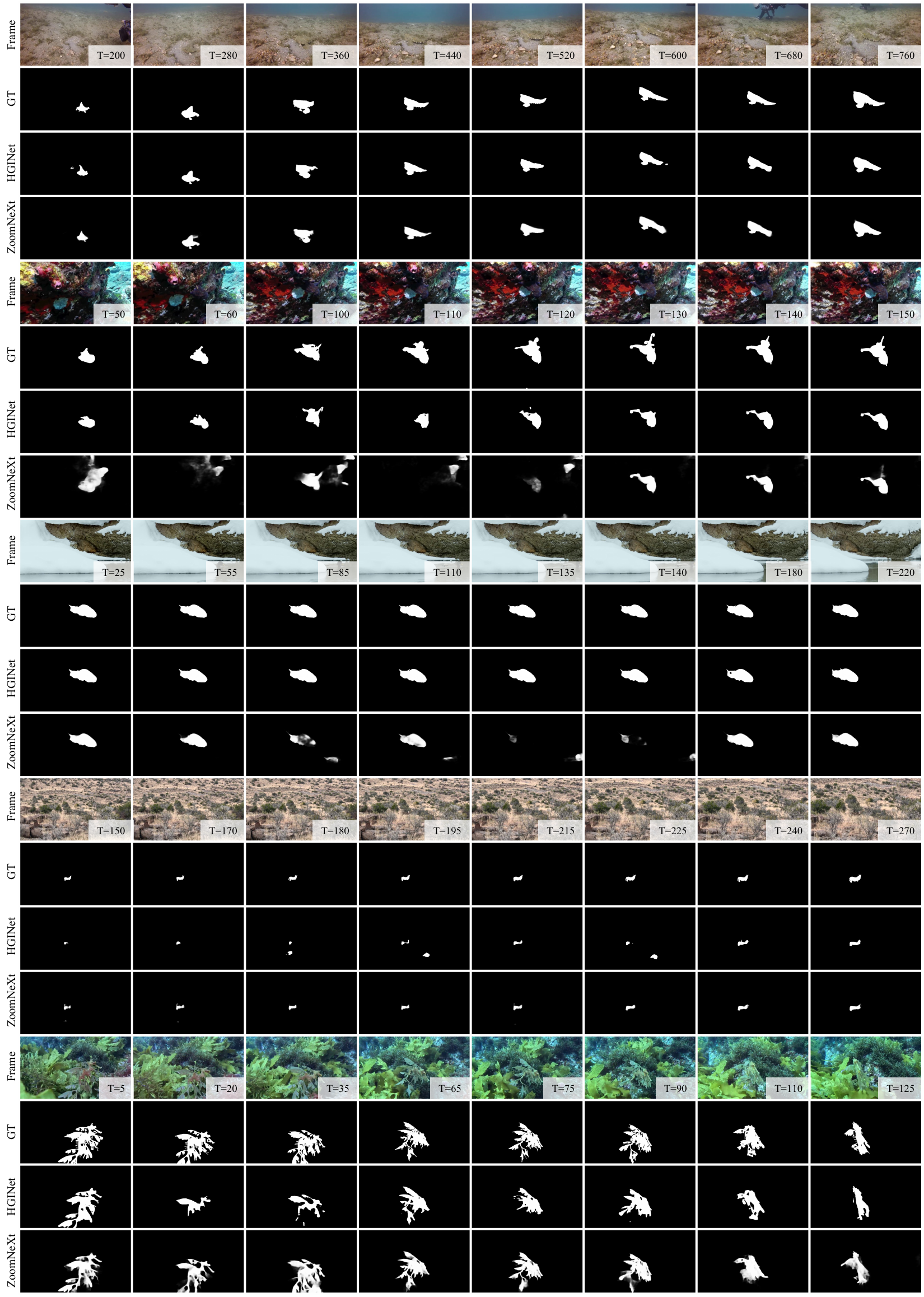}
\caption{Visual comparison with state-of-the-art methods on CAMotion. Please zoom in for details.}
\label{fig:qualitative_appendix}
\end{figure*}


\begin{figure*}[t]
    \centering   
    \begin{minipage}[t]{0.469\linewidth}
        \centering
        \includegraphics[width=\linewidth]{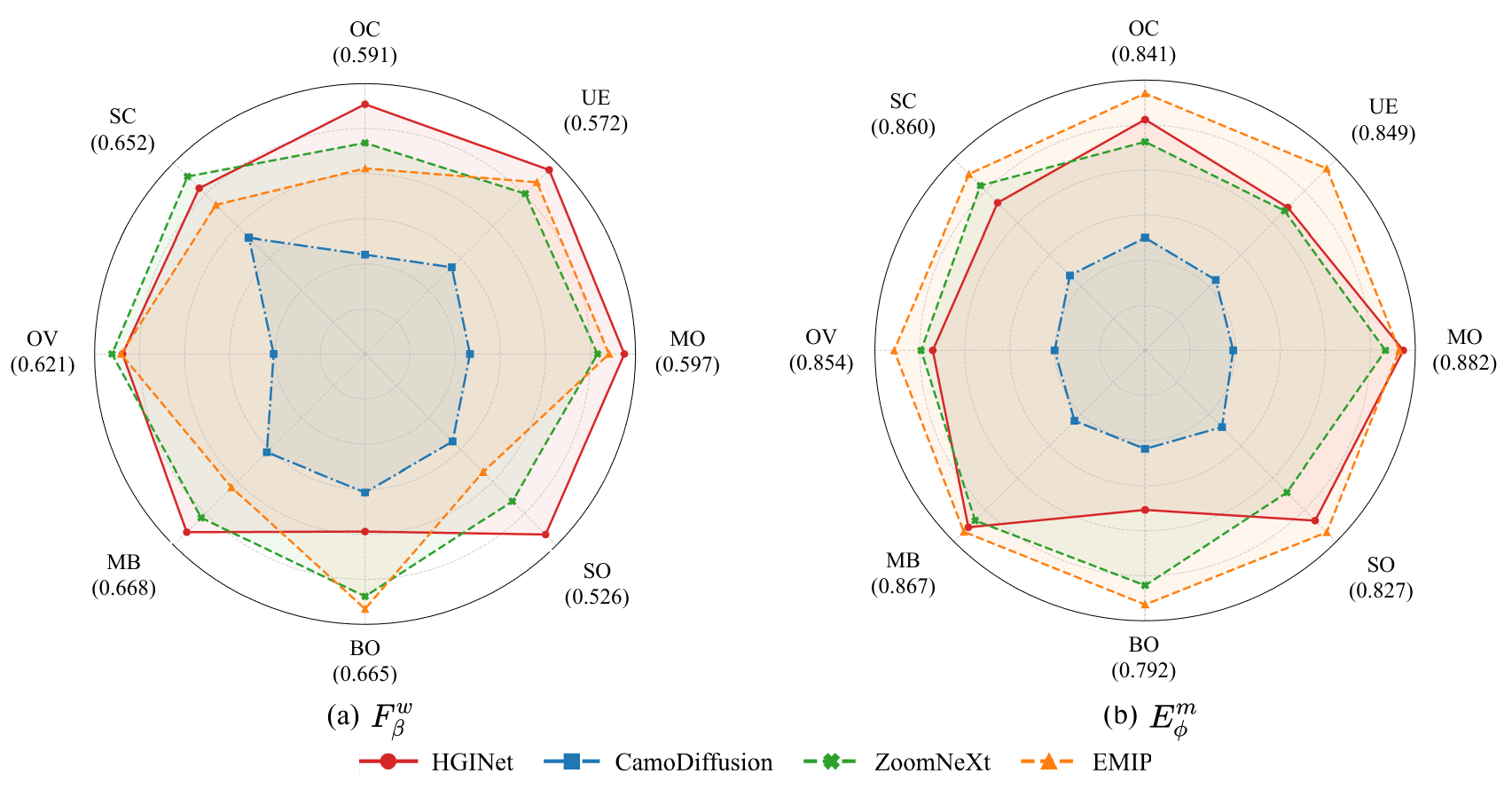}
        \caption{Visualization of SOTA method performances on different challenging attributes under (a) $F_{\beta}^{w}$ and (b) $E_\phi^m$.}
        \label{fig:attribute_performance_appendix}
    \end{minipage}
    \hfill
    \begin{minipage}[t]{0.523\linewidth}
        \centering
        \includegraphics[width=\linewidth]{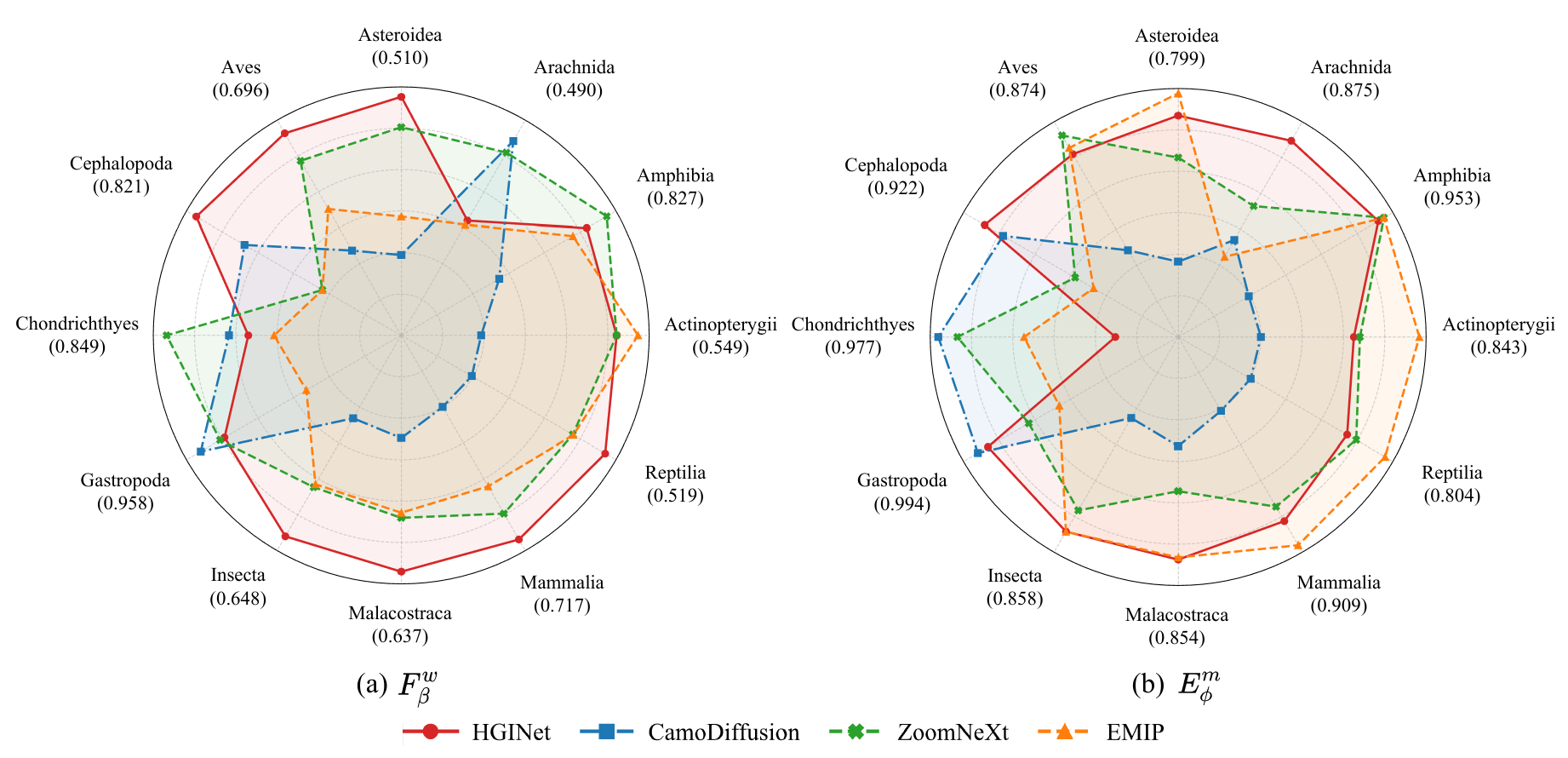}
        \caption{Visualization of SOTA method performances on different classes in terms of (a) $F_{\beta}^{w}$ and (b) $E_\phi^m$.}
        \label{fig:category_appendix}
    \end{minipage}
\end{figure*}

\begin{figure*}[!t]
\centering
\includegraphics[width=5.5in]{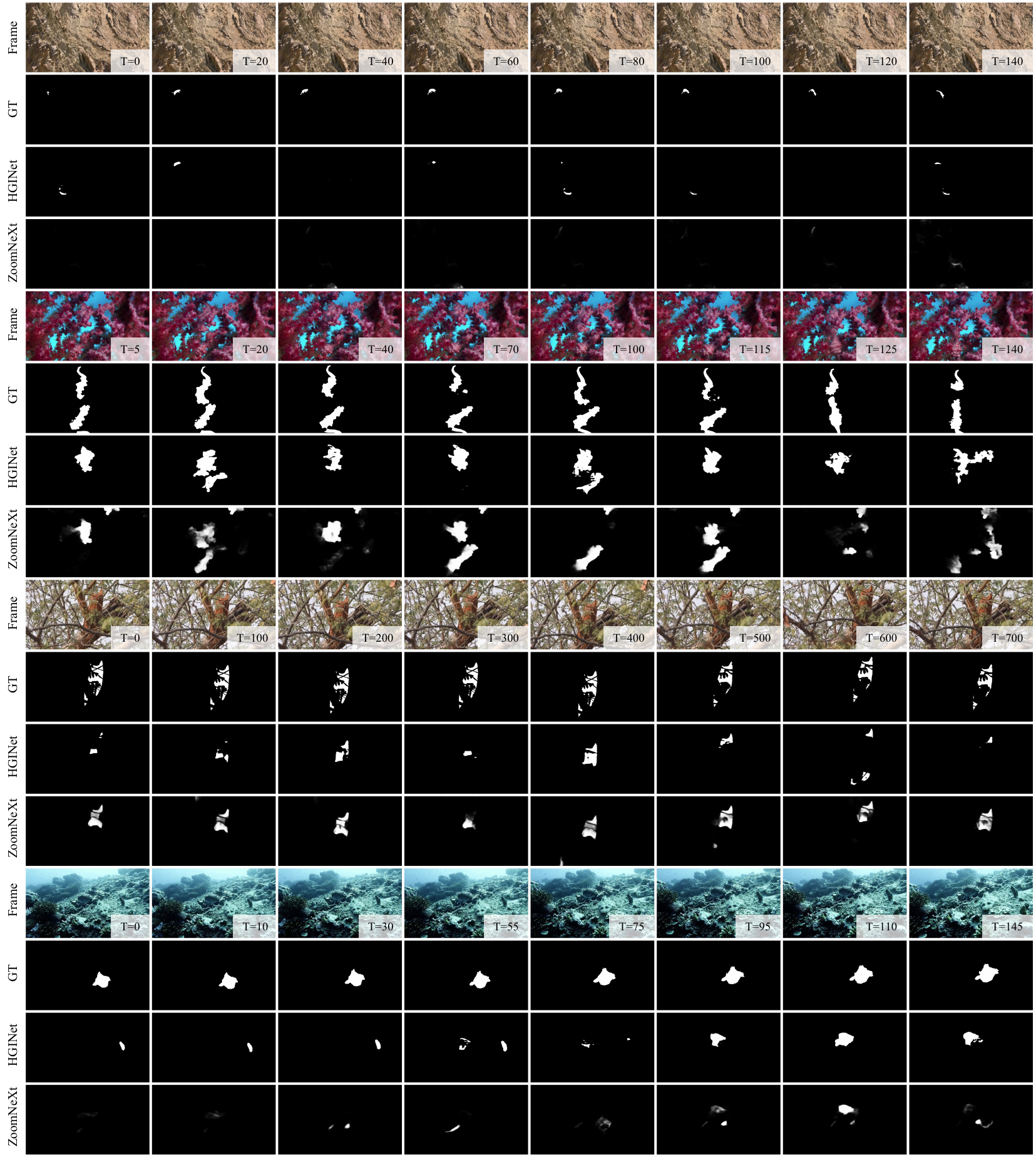}
\caption{Failure cases on both HGINet and ZoomNeXt. Please zoom in for details.}
\label{fig:failure_case_appendix}
\end{figure*}

\subsection{Qualitative Comparison}

As shown in Fig.~\ref{fig:qualitative_appendix}, we extend the visual comparison of HGINet \cite{DBLP:journals/tip/YaoSXWC24} and ZoomNeXt \cite{DBLP:journals/pami/PangZXZL24} in different challenging scenarios. Overall, both methods can identify the location and shapes of camouflaged objects in a subset of specific video frames. However, they are still distracted by highly confusing surrounding backgrounds, which affects the segmentation performance. For instance, when background elements (\textit{e.g.}, corals or aquatic plants) surrounding the camouflaged objects exhibit high similarity in both appearance and texture, they misguide the results and confuse the models into incorrect segmentation. Furthermore, models like ZoomNeXt may propagate distracting information across subsequent frames due to their limited discriminative ability. In contrast, for scenes with less confusing or distracting context, both methods demonstrate stronger discriminative capabilities.

\subsection{Dataset Analysis}
\label{ref_analysis}

\begin{figure*}[t]
    \centering
    \begin{minipage}[t]{0.496\linewidth}
        \vspace{0pt}
        \centering
        \includegraphics[width=\linewidth]{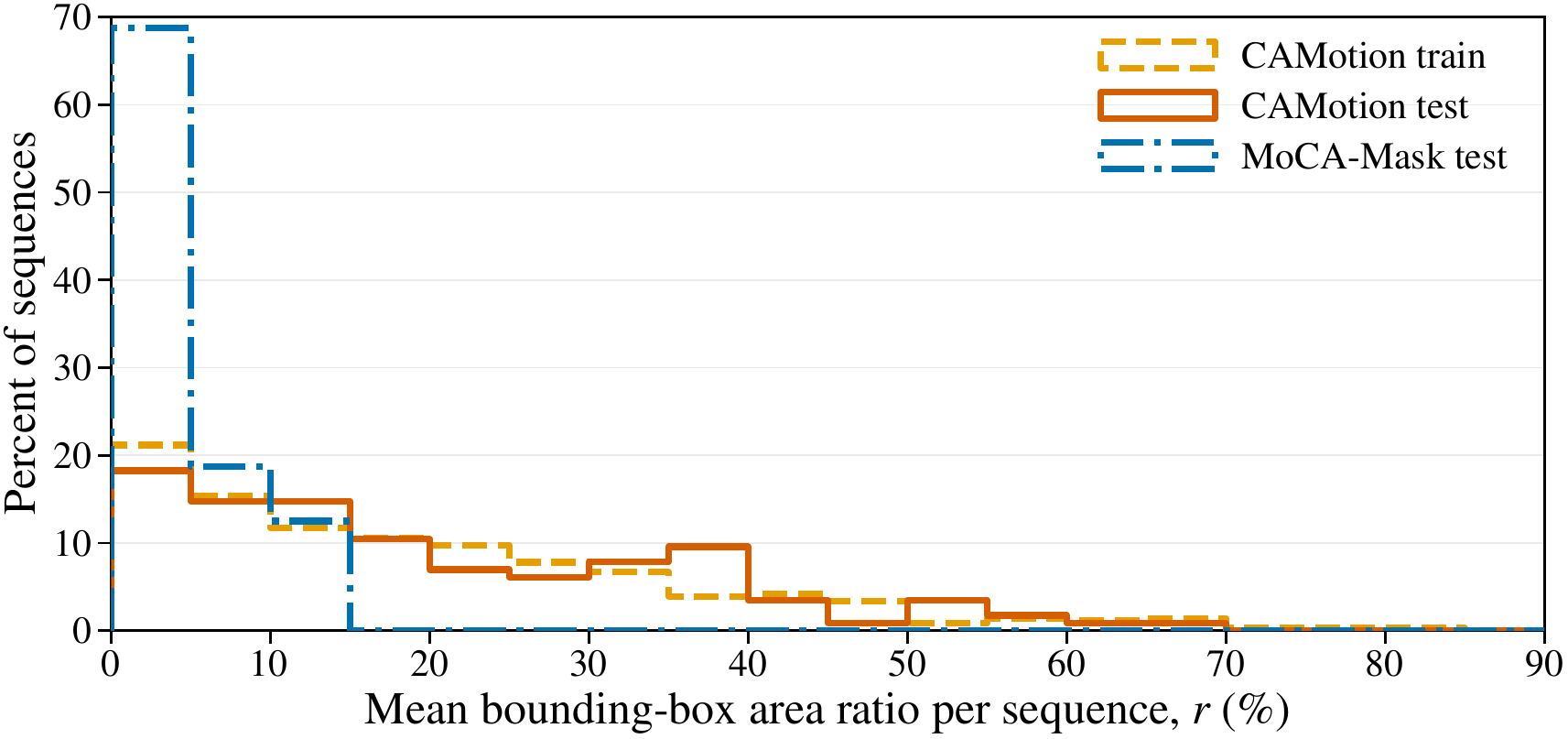}
        \caption{Histograms of sequence-level mean bounding-box area ratios for CAMotion and MoCA-Mask.}
        \label{fig:bbox_ratio_histogram}
    \end{minipage}
    \hfill
    \begin{minipage}[t]{0.49\linewidth}
        \vspace{0pt}
        \centering
        \includegraphics[width=\linewidth]{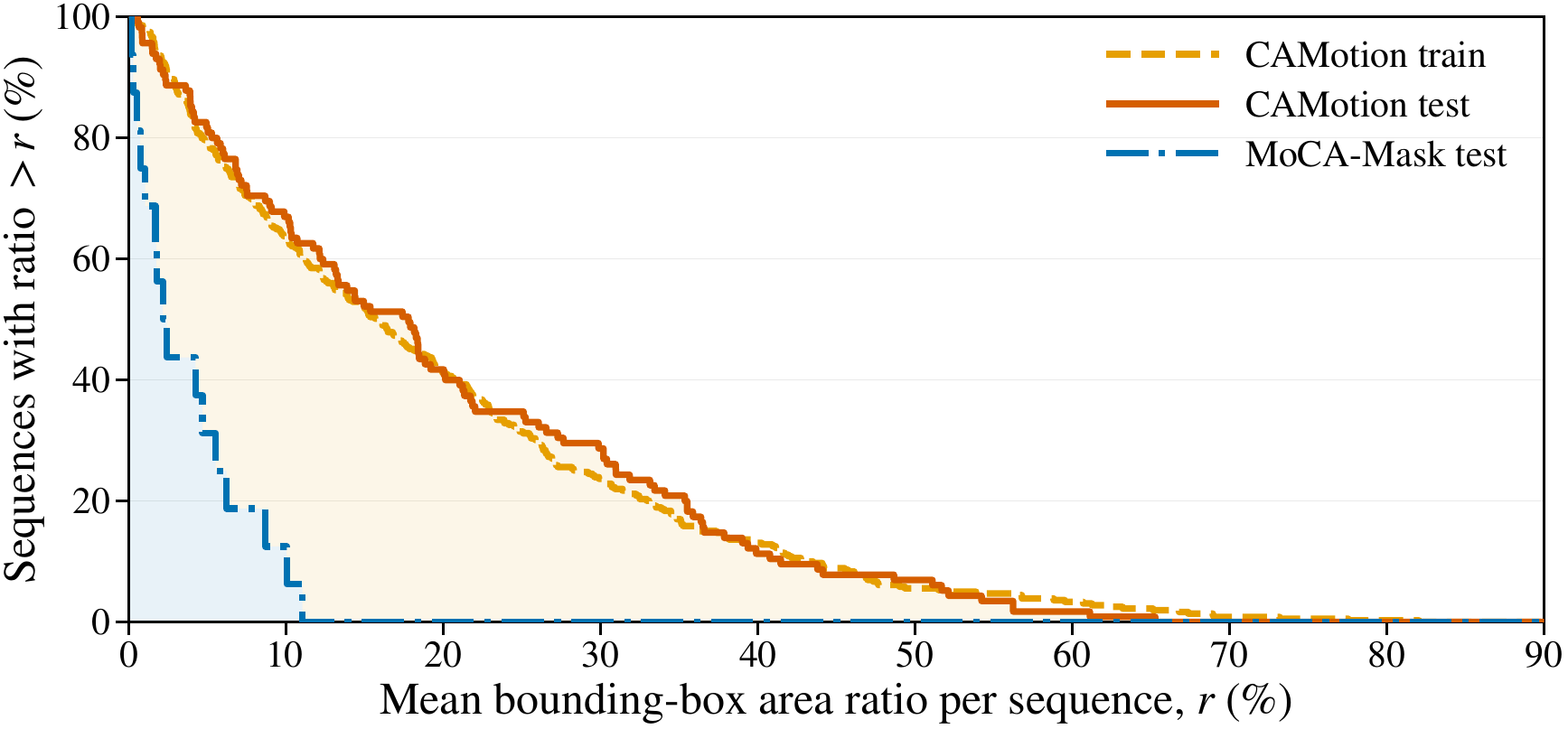}
        \caption{Empirical CCDFs of sequence-level mean bounding-box area ratios for CAMotion and MoCA-Mask.}
        \label{fig:bbox_ratio_ccdf}
    \end{minipage}
\end{figure*}

\begin{table*}[t]
  \scriptsize
  \centering
  \caption{Cross-dataset generalization of ZoomNeXt in terms of
  $S_{\alpha}$ and $\mathrm{mIoU}$, using 2,000 training frames and 500 testing frames per dataset. Self refers to training and testing on the same dataset, whereas Other refers to training on one dataset and testing on the other. Gap is the absolute difference between Self and Other. Bold indicates diagonal results.}
  \label{tab:cross_supp}
  \begin{tabular}{l|l|cc|cc|c}
    \toprule
    \textbf{Metric} &
    \diagbox{\textbf{Trained on}}{\textbf{Tested on}} &
    \textbf{CAMotion} &
    \textbf{MoCA-Mask} &
    \textbf{Self} &
    \textbf{Other} &
    \textbf{Gap} \\
    \midrule
    \multirow{2}{*}{$S_{\alpha}\uparrow$}
    & CAMotion
    & \textbf{0.749} & 0.629
    & 0.749 & 0.629 & 0.120 \\
    & MoCA-Mask
    & 0.697 & \textbf{0.607}
    & 0.607 & 0.697 & 0.090 \\
    \midrule
    \multirow{2}{*}{$\mathrm{mIoU}\uparrow$}
    & CAMotion
    & \textbf{0.496} & 0.260
    & 0.496 & 0.260 & 0.236 \\
    & MoCA-Mask
    & 0.403 & \textbf{0.206}
    & 0.206 & 0.403 & 0.197 \\
    \bottomrule
  \end{tabular}
\end{table*}

\begin{table*}[t]
\scriptsize
\centering
\caption{Ablation study of optical flow and depth cues on CAMotion.}
\label{tab:depth_ablation}
\begin{tabular}{lc@{\hspace{15pt}}cccccc}
\toprule
\multirow[c]{2}{*}[-0.8ex]{\textbf{Method}} &
\multirow[c]{2}{*}[-0.8ex]{\textbf{Input}} &
\multicolumn{6}{c}{\textbf{CAMotion (6,775)}} \\
\cmidrule(lr){3-8}
& &
$S_{\alpha}\uparrow$ &
$F_{\beta}^{w}\uparrow$ &
$E_{\phi}^{m}\uparrow$ &
$\mathit{M}\downarrow$ &
$\mathrm{mDic}\uparrow$ &
$\mathrm{mIoU}\uparrow$ \\
\midrule
\multirow{2}{*}{DaCOD~\cite{DBLP:conf/mm/WangYYWCZ23}}
& RGB-D    & 0.759 & 0.569 & 0.848 & 0.035 & 0.611 & 0.498 \\
& RGB-only & 0.728 & 0.521 & 0.826 & 0.039 & 0.562 & 0.447 \\
\midrule
\multirow{2}{*}{PopNet~\cite{wu2023source}}
& RGB-D    & 0.709 & 0.495 & 0.769 & 0.041 & 0.521 & 0.426 \\
& RGB-only & 0.702 & 0.475 & 0.781 & 0.044 & 0.510 & 0.408 \\
\midrule
\multirow{2}{*}{EMIP~\cite{DBLP:journals/tip/ZhangXJWFZ25}}
& RGB w/ optical flow & 0.761 & 0.583 & 0.866 & 0.035 & 0.617 & 0.506 \\
& RGB-only & 0.756 & 0.576 & 0.861 & 0.035 & 0.610 & 0.499 \\
\bottomrule
\end{tabular}
\end{table*}

\textbf{Cross-dataset generalization.}
Since our goal is to investigate the generalization capability and difficulty of CAMotion relative to MoCA-Mask while considering the diversity of species and scenes provided by each dataset, we therefore use the full training sets in the original experiment to retain this diversity, as shown in Table.~\ref{table4}. However, we need to control the number of training samples for a fair cross-dataset comparison, following~\cite{DBLP:conf/cvpr/TorralbaE11}. Therefore, we conduct an additional cross-dataset evaluation with the same training and testing subset sizes for CAMotion and MoCA-Mask. For both datasets, we randomly sample 400 five-frame clips from the original training set and 50 ten-frame clips from the original testing set, yielding 2,000 training frames and 500 testing frames. Each clip contains frames from a single video sequence, and the sampled clips do not share frames. For a fair comparison, we use the video version of ZoomNeXt as the base model.

As shown in Table~\ref{tab:cross_supp}, under the same training sample size, the model trained on CAMotion achieves higher $S_{\alpha}$ and $\mathrm{mIoU}$ scores on both testing subsets. Specifically, on the MoCA-Mask testing subset, it achieves $S_{\alpha}$ and $\mathrm{mIoU}$ scores of 0.629 and 0.260, compared with 0.607 and 0.206 for the model trained on MoCA-Mask, respectively. These results support the generalization advantage of CAMotion over MoCA-Mask even when the number of training samples is limited.

\textbf{Attribute-based performances.}
Fig.~\ref{fig:attribute_performance_appendix} illustrates the performance of HGINet \cite{DBLP:journals/tip/YaoSXWC24}, CamoDiffusion \cite{DBLP:journals/pami/SunCLSLJ25}, ZoomNeXt \cite{DBLP:journals/pami/PangZXZL24}, and EMIP \cite{DBLP:journals/tip/ZhangXJWFZ25} across eight challenging attributes in terms of $F_{\beta}^{w}$ and $E_\phi^m$. Consistent with the observations in the main text, the four methods exhibit relatively consistent performance trends across these two metrics. Notably, we can observe that the sequences involving small object (SO), uncertainty edge (UE) and occlusion (OC) are significantly more difficult. In contrast, sequences characterized by shape complexity (SC) and motion blur (MB) tend to yield relatively better performance. Furthermore, an evident difference is observed between big objects (BO) and small objects (SO). Attributed to the intrinsic properties of the evaluation metrics, big objects typically result in lower $E_\phi^m$ and higher $F_{\beta}^{w}$ scores, whereas small objects exhibit the opposite behavior.


\textbf{Class-based performances.}
To further evaluate the models across biological classes, Fig.~\ref{fig:category_appendix} presents the results of HGINet \cite{DBLP:journals/tip/YaoSXWC24}, CamoDiffusion \cite{DBLP:journals/pami/SunCLSLJ25}, ZoomNeXt \cite{DBLP:journals/pami/PangZXZL24}, and EMIP \cite{DBLP:journals/tip/ZhangXJWFZ25} evaluated using the $F_{\beta}^{w}$ and $E_\phi^m$ metrics on diverse species classes. Consistent with the observations in the main text, these methods demonstrate relatively stronger performance on \textit{Amphibia}, \textit{Cephalopoda}, \textit{Chondrichthyes}, and \textit{Gastropoda} and perform poorly on classes such as \textit{Actinopterygii} and \textit{Reptilia}. Still, CamoDiffusion and EMIP exhibit significant performance fluctuations across different classes on these two metrics.

\begin{figure}[t]
    \centering
    \includegraphics[width=\linewidth]{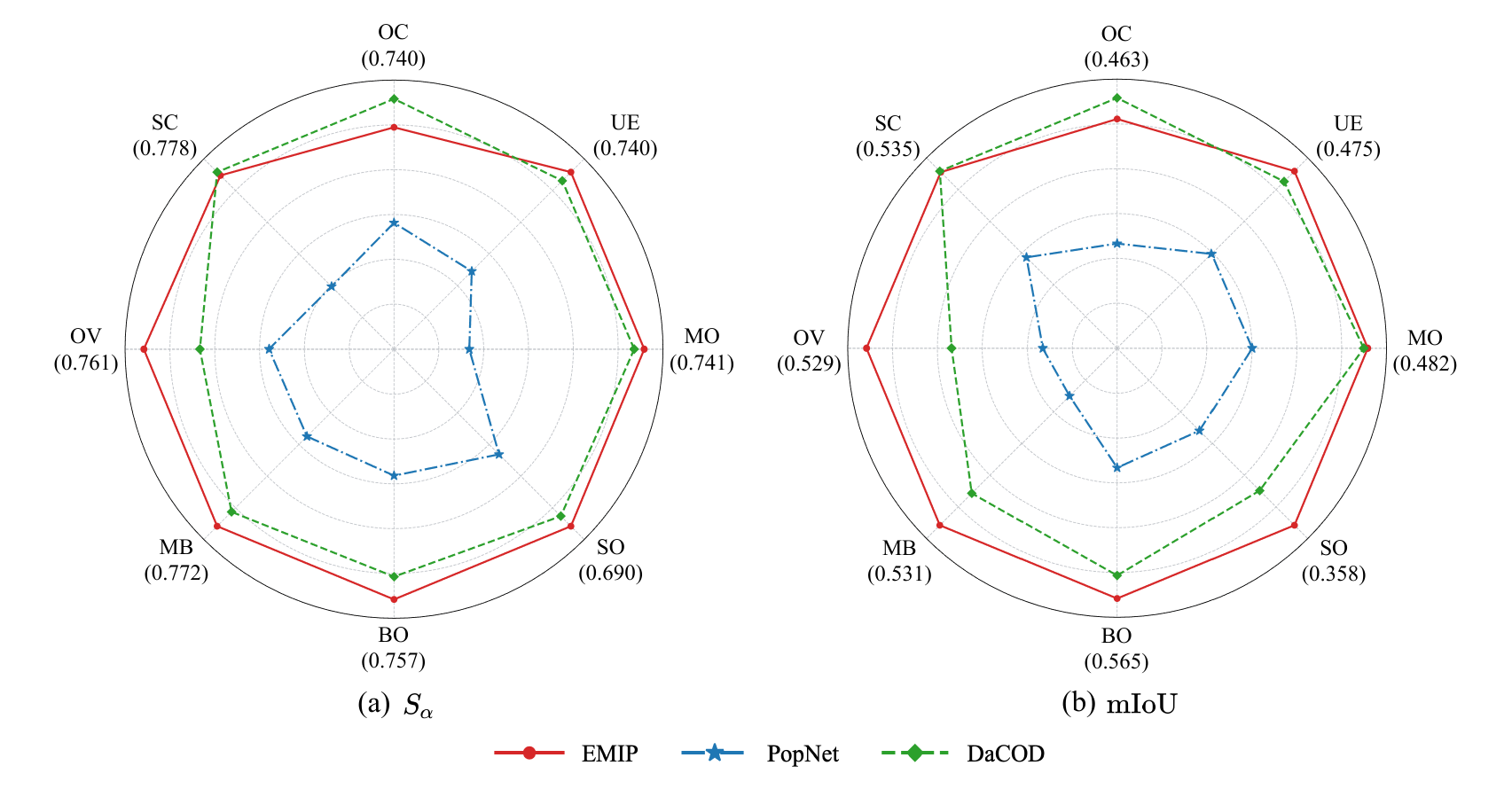}
    \caption{Performance of EMIP, PopNet, and DaCOD across different challenging attributes in terms of (a) $S_{\alpha}$ and (b) $\mathrm{mIoU}$.}
    \label{fig:modality_attribute_analysis}
\end{figure}

\begin{table*}[t]
    \footnotesize
    \centering
    \setlength{\tabcolsep}{3.5pt}
    \caption{Comparison of SAM-based methods on CAMotion under different first-frame prompt settings.}
    \label{tab:sam_prompt_analysis}
    \begin{tabular}{lcc@{\hspace{10pt}}cccccc}
    \toprule
    \multirow{2}{*}[-0.8ex]{\textbf{Method}} &
    \multirow{2}{*}[-0.8ex]{\makecell{\textbf{First-frame}\\\textbf{Prompt}}} &
    \multirow{2}{*}[-0.8ex]{\makecell{\textbf{\#Evaluated}\\\textbf{Frames}}} &
    \multicolumn{6}{c}{\textbf{CAMotion}} \\
    \cmidrule(lr){4-9}
    & & &
    $S_{\alpha}\uparrow$ &
    $F_{\beta}^{w}\uparrow$ &
    $E_{\phi}^{m}\uparrow$ &
    $\mathit{M}\downarrow$ &
    $\mathrm{mDic}\uparrow$ &
    $\mathrm{mIoU}\uparrow$ \\
    \midrule
    \multicolumn{9}{c}{\emph{Without first-frame prompts}} \\
    \midrule
    SAM2~\cite{DBLP:conf/iclr/RaviGHHR0KRRGMP25}
    & None & 6,775
    & 0.463 & 0.004 & 0.256 & 0.084 & 0.004 & 0.003 \\
    SAM-PM~\cite{DBLP:conf/cvpr/MeeranTM22}
    & None & 6,775
    & 0.741 & 0.550 & 0.802 & 0.047 & 0.589 & 0.488 \\
    CamSAM2~\cite{zhou2025camsam}
    & None & 6,775
    & 0.626 & 0.378 & 0.701 & 0.075 & 0.393 & 0.328 \\
    \midrule
    \multicolumn{9}{c}{\emph{With pseudo first-frame prompts}} \\
    \midrule
    HGINet~\cite{DBLP:journals/tip/YaoSXWC24} + SAM2
    & Pseudo mask & 6,775
    & 0.758 & 0.601 & 0.799 & 0.031 & 0.623 & 0.527 \\
    HGINet~\cite{DBLP:journals/tip/YaoSXWC24} + CamSAM2
    & Pseudo mask & 6,775
    & 0.775 & 0.646 & 0.833 & 0.035 & 0.669 & 0.570 \\
    \bottomrule
    \end{tabular}
\end{table*}

\textbf{Scale distribution analysis.} To deeper analyse the scale distribution, we calculate the bounding-box-to-image area ratio for each annotated frame and average these ratios within each video sequence. The resulting histograms and empirical complementary cumulative distributions (CCDFs) are presented in Figs.~\ref{fig:bbox_ratio_histogram} and \ref{fig:bbox_ratio_ccdf}, respectively. As shown in these figures, the CAMotion testing set exhibits a broader scale distribution than the MoCA-Mask testing set, allowing model performance to be evaluated across more diverse object scales. Moreover, the closely aligned distributions of the CAMotion training and testing sets provide a consistent basis for training and evaluation.

\textbf{Optical flow and depth analysis.} To investigate the performance of methods using optical flow and depth cues under different challenging attributes, we evaluate optical-flow-based EMIP~\cite{DBLP:journals/tip/ZhangXJWFZ25} and two RGB-D COD methods, \emph{i.e.}, PopNet~\cite{wu2023source} and DaCOD~\cite{DBLP:conf/mm/WangYYWCZ23}, across eight challenging attributes on CAMotion. The optical flow and depth maps are generated using GMFlow~\cite{DBLP:journals/pami/XuZCRYTG23} and Depth Anything V2~\cite{DBLP:conf/nips/YangKH0XFZ24}, respectively. As shown in Fig.~\ref{fig:modality_attribute_analysis}, the optical-flow-based method EMIP performs worse than the depth-based method DaCOD on occlusion and shape complexity. Meanwhile, DaCOD performs worse than the optical-flow-based method EMIP in other attributes, particularly out-of-view. These results reveal that relying solely on either depth or optical flow cues is insufficient for the VCOD task under these attributes.

\textbf{Ablation analysis on auxiliary cues.} To further investigate the effectiveness of these auxiliary cues, we conduct ablation experiments with and without optical flow for EMIP and with and without depth input for DaCOD and PopNet. As shown in Table~\ref{tab:depth_ablation}, incorporating these cues could improve the segmentation performance. Specifically, compared with its RGB-only variant, DaCOD achieves relative performance gains of 4.3\%, 9.2\%, 2.7\%, 10.3\%, 8.7\%, and 11.4\% in terms of $S_{\alpha}$, $F_{\beta}^{w}$, $E_{\phi}^{m}$, $\mathit{M}$, $\mathrm{mDic}$, and $\mathrm{mIoU}$, respectively. With optical flow, EMIP obtains higher scores on most metrics, with $\mathrm{mIoU}$ increasing from 0.499 to 0.506. These results confirm that either estimated optical flow or depth maps can provide complementary information that improves camouflaged object segmentation accuracy.

\textbf{Comparison of SAM-based methods.} Since we follow the VCOD setting~\cite{DBLP:conf/cvpr/ChengXFZHDG22} and evaluate methods on CAMotion without first-frame prompts, accordingly, we report the results of SAM2~\cite{DBLP:conf/iclr/RaviGHHR0KRRGMP25} and CamSAM2~\cite{zhou2025camsam} under this setting in Table~\ref{tab:table3}. Since these methods rely on prompts to locate the target objects, the reported scores reflect their performance without first-frame guidance rather than under their original prompt settings. To further investigate the influence of first-frame prompts, we evaluate SAM2 and CamSAM2 using pseudo masks predicted by HGINet~\cite{DBLP:journals/tip/YaoSXWC24} as first-frame prompts. As shown in Table~\ref{tab:sam_prompt_analysis}, both methods achieve better results across all six metrics with the predicted prompts. Specifically, the $\mathrm{mIoU}$ scores of SAM2 and CamSAM2 increase from 0.003 and 0.328 to 0.527 and 0.570, respectively.

\textbf{Failure cases.} 
As shown in Fig.~\ref{fig:failure_case_appendix}, we present more failure cases of HGINet \cite{DBLP:journals/tip/YaoSXWC24} and ZoomNeXt \cite{DBLP:journals/pami/PangZXZL24} in challenging scenarios. In the first sequence (Rows 1–4), both methods struggle to locate and identify the small camouflaged object due to the highly confusing similarity between the object and its surrounding background. Similar challenges are observed in the sequence presented in Rows 13–16. In the second sequence (Rows 5–8), these two methods tend to focus on visually salient regions, indicating that they may be misled by the salient regions and resulting in incorrect segmentation. Rows 9–12 show that the segmentation results remain fragmented and imprecise due to the highly similar color and texture patterns between the object and its surrounding background, although the camouflaged object is partially detected. Overall, these results highlight the diversity and difficulty of our proposed CAMotion dataset, reinforcing its value as a rigorous benchmark for advancing research in video camouflaged object detection.

\bmhead{Acknowledgements} This work is supported by the National Natural Science Foundation of China (No. 62402055, 62322216, 62025604, 62302480), the Shenzhen Science and Technology Program (CJGJZD20240729141505007).

\bmhead{Data Availability} The CAMotion dataset, well-trained checkpoints, generated depth map and optical flow, and related materials are available at \url{https://www.camotion.focuslab.net.cn}.

\bibliographystyle{spmpsci}      
\bibliography{bibliography}   

\end{document}